\documentclass{article} %
\usepackage{iclr2026_conference, times}

\usepackage[utf8]{inputenc} %
\usepackage[T1]{fontenc}    %
\usepackage{hyperref}       %
\usepackage{url}            %
\usepackage{booktabs}       %
\usepackage{amsfonts}       %
\usepackage{nicefrac}       %
\usepackage{microtype}      %
\usepackage{xcolor}         %

\usepackage{microtype}
\usepackage{subcaption}
\usepackage{graphicx}
\usepackage{booktabs} %

\usepackage{xspace}
\usepackage{multicol}
\usepackage{multirow}
\usepackage{pifont}%
\newcommand{\cmark}{\ding{51}}%
\newcommand{\xmark}{\ding{55}}%

\usepackage{hyperref}
\usepackage{xurl} %
\usepackage{lipsum}
\usepackage{nicefrac}
\usepackage{makecell}
\usepackage{float}
\usepackage{enumerate}

\usepackage{amsmath}
\usepackage{amssymb}
\usepackage{mathtools}
\usepackage{amsthm}
\usepackage{wasysym}

\usepackage[linesnumbered,ruled,vlined]{algorithm2e}
\SetKwComment{Comment}{$\triangleright$\ }{}
\SetCommentSty{mycommentsty} %

\usepackage{siunitx}
\usepackage{wrapfig}

\usepackage[capitalize,noabbrev]{cleveref}

\theoremstyle{plain}

\theoremstyle{definition}

\theoremstyle{remark}

\usepackage[textsize=tiny]{todonotes}

\usepackage{acronym}
\usepackage[abbreviations]{foreign}

\usepackage{color,colortbl}
\usepackage{adjustbox}
\usepackage{xcolor}

\definecolor{mygray}{gray}{.9}
\definecolor{ggray}{RGB}{127,127,127}
\definecolor{reda}{RGB}{192,0,0}
\definecolor{redb}{RGB}{217,148,143}
\definecolor{myyellow}{RGB}{190,144,0}
\definecolor{mygreen}{RGB}{80,100,40}
\definecolor{myblue}{RGB}{30,90,100}

\definecolor{darkgreen}{rgb}{0,0.5,0}
\definecolor{azureblue}{rgb}{0,0.5,1}
\definecolor{darkgreen}{rgb}{1,0,0}
\definecolor{color1}{HTML}{006EB8}
\definecolor{color2}{HTML}{009B55}
\definecolor{color3}{HTML}{00A99A}
\definecolor{color4}{HTML}{3C8031}
\definecolor{color5}{HTML}{006795}
\definecolor{color6}{HTML}{00AEB3}
\definecolor{mygray}{gray}{0.93}
\definecolor{mygreen}{HTML}{3FBC9D}
\definecolor{arsenic}{rgb}{0.23, 0.27, 0.29}

\newcommand{\methodshort}[1]{PCB-Merging}

\DeclareMathOperator*{\argmax}{arg\,max}

\usepackage{paralist}
\usepackage[shortlabels]{enumitem} %

\newcommand{\twinm}{\textsc{Twin-Merging}\xspace}
\newcommand{\tsvm}{\textsc{TSV-M}\xspace}
\newcommand{\consensusm}{\textsc{Consensus}\xspace}
\newcommand{\tam}{\textsc{TA}\xspace}
\newcommand{\ties}{\textsc{TIES-Merging}\xspace}
\newcommand{\adam}{\textsc{AdaMerging}\xspace}
\newcommand{\emr}{\textsc{EMR-Merging}\xspace}
\newcommand{\regmean}{\textsc{RegMean}\xspace}
\newcommand{\fisher}{\textsc{Fisher Merging}\xspace}
\newcommand{\wemoe}{\textsc{WEMoE}\xspace}
\newcommand{\pcbm}{\textsc{PCB-Merging}\xspace}
\newcommand{\wudim}{\textsc{WUDI-Merging}\xspace}
\newcommand{\surgery}{\textsc{Surgery}\xspace}
\newcommand{\sys}{\textsc{FlexMerge}\xspace}
\newcommand{\dare}{\textsc{DARE}\xspace}

\newcommand{\channelm}{\textsc{Channel Merging}\xspace}
\newcommand{\linesm}{LiNeS\xspace}

\newcommand{\btheta}{\boldsymbol{\theta}}
\newcommand{\bthetah}{\boldsymbol{\hat{\theta}}}
\newcommand{\mask}{\boldsymbol{m}}
\newcommand{\tv}{\boldsymbol{\tau}}
\newcommand{\tvmtl}{\boldsymbol{\tau}_{\text{uni}}}
\acrodef{DL}{decentralized learning}
\acrodef{ML}{machine learning}
\acrodef{D-PSGD}{decentralized parallel stochastic gradient descent}
\acrodef{FL}{federated learning}
\acrodef{FI}{federated inference}
\acrodef{FU}{federated unlearning}
\acrodef{SGD}{stochastic gradient descent}
\acrodef{IID}{independent and identically distributed}
\acrodef{non-IID}{non independent and identically distributed}
\acrodef{RMSE}{root mean square error}
\acrodef{RMW}{random model walk}
\acrodef{GL}{gossip learning}
\acrodef{DWT}{discrete wavelet transform}
\acrodef{LAN}{local area network}
\acrodef{WAN}{wide area network}
\acrodef{NN}{neural network}
\acrodef{KD}{knowledge distillation}
\acrodef{DD}{dataset distillation}
\acrodef{GDPR}{General Data Protection Regulation}
\acrodef{SOTA}{state-of-the-art}
\acrodef{CCPA}{California Consumer Privacy Act}
\acrodef{MIA}{membership inference attack}
\acrodef{SGA}{stochastic gradient ascent}
\acrodef{SGD}{stochastic gradient descent}
\acrodef{MU}{machine unlearning}
\acrodef{NLP}{natural language processing}
\acrodef{PEFT}{parameter efficient fine-tuning}
\acrodef{FFT}{full-parameter fine-tuning}
\acrodef{FT}{fine-tuning}
\acrodef{MTL}{multi-task learning}
\acrodef{MoE}{mixture-of-experts}
\acrodef{OOD}{Out-of-Distribution}

\newcommand{\tzerob}{T0-3B\xspace}
\newcommand{\tfivebase}{T5-Base\xspace}
\newcommand{\tfivelarge}{T5-Large\xspace}
\newcommand{\vitb}{\mbox{ViT-B/32}\xspace}
\newcommand{\vitl}{\mbox{ViT-L/14}\xspace}
\newcommand{\beit}{\mbox{BEiT-3-base}\xspace}

\newcommand{\codeurl}{\url{https://github.com/sacs-epfl/flexmerge}}
\newboolean{showcomments}
\setboolean{showcomments}{false}
\ifthenelse{\boolean{showcomments}}
{ \newcommand{\mynote}[3]{
		\fbox{\bfseries\sffamily\scriptsize#1}
		{\small$\blacktriangleright$\textsf{\emph{\color{#3}{#2}}}$\blacktriangleleft$}}
	\newcommand{\zzz}[1]{{\setlength{\fboxsep}{2pt}\fcolorbox{black}{yellow}{\textsf{\emph{#1}}}}\xspace}}
{ \newcommand{\mynote}[3]{}
	\newcommand{\zzz}[1]{}}

\definecolor{BlueViolet}{RGB}{138,43,226}

\title{Navigating the Accuracy-Size Trade-Off with \\ Flexible Model Merging}

\author{Akash Dhasade$^{1}$\thanks{Work done during research visit to Carnegie Mellon University.}, Divyansh Jhunjhunwala$^2$, Milos Vujasinovic$^1$, Gauri Joshi$^2$, \\
\textbf{Anne-Marie Kermarrec$^1$} \\
$^1$EPFL, Switzerland \quad $^2$Carnegie Mellon University, USA \\
\texttt{akash.dhasade@epfl.ch} \quad \texttt{milos.vujasinovic@epfl.ch} \\
}

\iclrfinalcopy 
\begin{document}

\maketitle

\begin{abstract}
Model merging has emerged as an efficient method to combine multiple single-task fine-tuned models. 
The merged model can enjoy multi-task capabilities without expensive training. 
While promising, merging into a single model often suffers from an accuracy gap with respect to individual fine-tuned models.
On the other hand, deploying all individual fine-tuned models incurs high storage costs.
We propose \sys, a novel data-free model merging framework that: \emph{(a)} flexibly generates merged models of varying sizes, spanning the full spectrum from a single merged model to retaining all individual fine-tuned models; and \emph{(b)} supports multiple merging algorithms in a unified framework.
Using \sys, we systematically characterize the accuracy–size trade-off of different  algorithms. Our study reveals two key findings: first, even modestly larger merged models can yield steep accuracy gains (up to $13.5\%$ when just doubling the size); second, algorithm rankings are not consistent as size increases, with some methods overtaking others beyond the one-model regime. These results uncover a new design dimension for model merging: developing and comparing algorithms across the full spectrum of sizes rather than only at the single-model limit. Extensive experiments on vision and NLP benchmarks, with up to $30$ tasks, confirm the generality and practicality of \sys.\footnote{Our code is available at \codeurl{}}

\end{abstract}

\section{Introduction}
\label{sec:introduction}
In recent years, the pre-training followed by fine-tuning paradigm has become the leading approach in both \ac{NLP} and computer vision, showcasing remarkable success on a wide range of tasks~\citep{devlin2018bert,dodge2020fine,dosovitskiy2021an,bommasani2021opportunities}.
Pre-trained models (PTMs), which learn generalized features from large-scale datasets, serve as powerful starting points, enabling fine-tuning to achieve superior performance on downstream tasks with less labeled data. 
This has led to an exponential growth in the number of fine-tuned models driven further by the availability of open-source repositories~\citep{torchvision2016,wolf2019huggingface}. 
However, \textit{deploying individual fine-tuned models} for specific tasks incurs high storage and deployment costs. 
The alternative is \Ac{MTL}, which aims to jointly train \textit{a single model} across multiple tasks~\citep{vandenhende2021multi,sanh2022multitask}. 
But \ac{MTL} comes with its own drawbacks, such as significant computational overhead and the need to simultaneously access the data from all tasks, which might be infeasible due to privacy constraints~\citep{jin2023dataless}. 

To mitigate these limitations, model merging has emerged as a promising solution, allowing the combination of multiple fine-tuned models into a \emph{single model} without access to training data. 
To this end, several model merging methods have been proposed~
\citep{gargiulo2025task,huang2024emrmerging,pmlr-v235-yang24t,NEURIPS2023_1644c9af,ilharco2023editing,matena2022merging}.
However, a single model is often unable to perfectly resolve parameter conflicts between tasks, leaving an accuracy gap with respect to the individual fine-tuned models~\citep{Zhang_Liu_Ding_Ou_Yu_Zhuang_2025,huang2024emrmerging}. 
This gap becomes more significant as a higher number of models are merged~\citep{NEURIPS2023_1644c9af,ilharco2023editing}. 
To mitigate this issue, some methods leverage additional data to facilitate merging~\citep{lu2024twinmerging,pmlr-v235-yang24t,tang2024merging,yang2024adamerging}.
Yet, the data-dependency might be difficult to meet in practice due to privacy constraints or proprietary restrictions, leading to a growing focus on data-free model merging techniques~\citep{gargiulo2025task,huang2024emrmerging,du2024parameter,yu2024language,NEURIPS2023_1644c9af}.
Nevertheless, in the absence of data, the accuracy gap remains significant, highlighting the need for novel solutions. 

 \begin{figure}[t!]
  \centering
  \begin{subfigure}[b]{0.64\textwidth}
    \vspace{-10pt}
    \includegraphics[width=\linewidth]{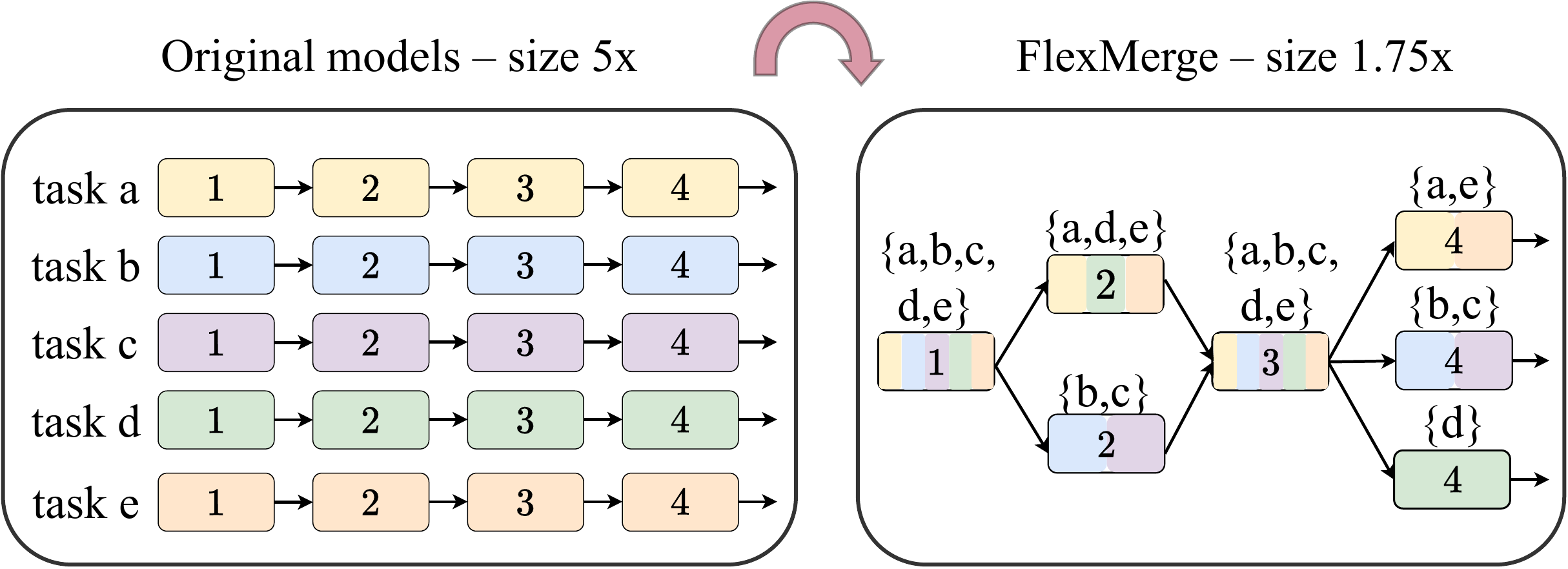}
    \caption{\sys illustration}
    \label{fig:sub1}
  \end{subfigure}
  \hfill
  \begin{subfigure}[b]{0.35\textwidth}
    \vspace{-10pt}
    \includegraphics[]{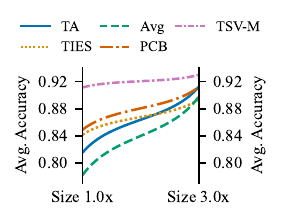}
    \caption{\sys on $8$ tasks (\vitl)}
    \label{fig:sub2}
  \end{subfigure}
  \caption{(a) Fine-tuned models are sequences of blocks. \sys iteratively merges block pairs until reaching the desired size (\eg size $1.75\times$). (b) Algorithm rankings change as size is increased.}
  \label{fig:intro_fig}
\end{figure}

We argue that an effective solution to this challenge is to go beyond the conventional one model approach, and merge into model(s) of bigger sizes.
Merging multiple fine-tuned models naturally presents a trade-off between maintaining accuracy and achieving model compactness, dictated by the size of the merged model. 
This trade-off spans a spectrum: at one extreme, retaining all individual fine-tuned models for each task achieves maximal accuracy but at the cost of larger overall size; at the other, fully merging all tasks into a single model minimizes storage size but sacrifices accuracy. 
Despite this clear trade-off, a systematic investigation of the accuracy-size relationship in model merging has been lacking.
In this light, we pose two key research questions: \emph{(RQ1) How can we derive merged models across the full range of model sizes in a data-free manner?} and \emph{(RQ2) What is the nature of the accuracy-size trade-off exhibited by different data-free merging algorithms?}

\begin{wrapfigure}[27]{r}{0.3\textwidth} %
	\centering
	\vspace{-10 pt}
	\includegraphics[]{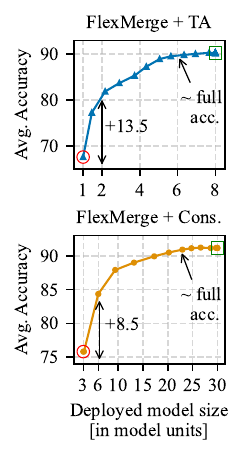}
	\caption{\sys enables large accuracy gains when just doubling the deployed model size and attains full accuracy well before the maximum size.}
	\label{fig:motivation}
\end{wrapfigure}

In response to (RQ1), we propose \sys, a flexible framework that enables \textit{data-free} fusion into model(s) of any desired size. 
At its core, \sys treats each fine-tuned model as composed of sequential blocks, as illustrated in \Cref{fig:intro_fig}(a), whose granularity can be controlled (\eg a transformer block, a few layers, or even a single layer).
It then takes a bottom-up approach starting with all fine-tuned models with their respective blocks and greedily merging a pair of blocks with the highest cosine similarity in each merging iteration. 
This merging can leverage \textit{any} existing data-free merging method such as Task Arithmetic (TA)~\citep{ilharco2023editing}, \ties~\citep{NEURIPS2023_1644c9af}, \emr~\citep{huang2024emrmerging}, \tsvm~\citep{gargiulo2025task}, \etc., applied at the block-level.
With each merging iteration, the size of the deployed model is reduced, and the process can be halted once the desired size is met. 
For instance, in \Cref{fig:intro_fig}(a), the merging is halted when the merged model is $1.75\times$ the size of a single fine-tuned model. 
The entire merging process in \sys needs no additional data or tuning, making \sys fully \textit{data-free}.

In response to (RQ2), we demonstrate with \sys that a range of data-free merging algorithms exhibit highly favorable accuracy-size trade-offs.
Remarkably, the accuracy-size trade-off is characterized by steep gains in accuracy for even modestly bigger merged models beyond one model, followed by steady improvements, reaching near fine-tuning accuracy well before the maximum size.
To illustrate this in practice, \Cref{fig:motivation} charts the merged model accuracy versus deployed size for 8 tasks (top) and 30 tasks (bottom) using the \vitb model, with \tam~\citep{ilharco2023editing} and \consensusm~\citep{wang2024localizing} as the respective merging methods.
$\Circle$ and $\square$ annotate the accuracy at both ends of the spectrum \ie lowest fused size and retaining all fine-tuned models respectively.
\sys + \tam gains $13.5\%$ in average accuracy when going from $1\times$ to $2\times$ while \sys + \consensusm gains $8.5\%$ when doubling the size from approximately $3\times$ to $6\times$.
We note that \consensusm requires storing masks and the pre-trained parameters alongside the unified parameters~\citep{wang2024localizing}, resulting in the lowest possible size of $\approx 3\times$ for $30$ tasks.  
We observe that the steep rise is followed by relatively slower accuracy growth in the middle. 
Yet, a near fine-tuning accuracy is attained well before the maximum size.
For $8$ tasks, this is obtained around size $6\times$ and for $30$ tasks, around size $23.5\times$.
Secondly, we observe that algorithm rankings are not consistent even at modestly bigger sizes. 
\Cref{fig:intro_fig}(b) shows that vanilla averaging exceeds \ties while \tam attains the performance of \pcbm at size $3\times$ despite starting from a large gap at $1\times$.
\textit{Our findings open a new design dimension: encouraging algorithm development and comparison for sizes $> 1\times$ instead of restricting only to $1\times$.}

\textbf{Contributions.} To the best of our knowledge, we present the first study of model merging that:
\begin{itemize}[leftmargin=15pt]
\item Generates merged models across full spectrum of sizes, \textit{including non-integer sizes};

\item Supports a wide range of data-free merging algorithms, \textit{within a unified framework};

\item Provides a systematic characterization of the accuracy-size trade-off in data-free model merging, \textit{revealing general trends, highly favorable regions and inconsistency of algorithm rankings};

\item Demonstrates that larger merged sizes incur negligible inference-time overhead, \textit{enabled by our efficient implementation}.
\end{itemize}
We confirm our findings through extensive experiments spanning language and vision modalities, multiple model families, multi-modal datasets, using both \ac{FFT} and \ac{PEFT}, scaling up to $30$ tasks.

\section{Related Work}
\label{sec:related_work}

Initial studies on model merging focused on vanilla averaging as a way of combining models obtained from same or different training runs of a task into one higher performing model \citep{IzmailovPGVW18, DBLP:conf/iclr/GuptaSD20, wortsman2022model, DBLP:conf/nips/ChaCLCPLP21}. Vanilla averaging is also used in federated learning to merge different client models \citep{DBLP:conf/aistats/McMahanMRHA17, DBLP:journals/corr/KonecnyMRR16}. 
\citet{ilharco2023editing} introduced task vectors, representing the difference between fine-tuned and pre-trained models, enabling model combination through vector arithmetic.

\textbf{Data-based merging} methods leverage validation data to facilitate merging. 
Techniques like \fisher~\citep{matena2022merging} and \regmean~\citep{jin2023dataless} compute the Fisher Information and Gram matrices, respectively, for weighted averaging of model parameters. 
\surgery~\citep{pmlr-v235-yang24t} trains task-specific adapters to debias the representations produced by the merged model.
\adam~\citep{yang2024adamerging} introduces per-task, per-layer merging co-efficients, and proposes to learn these co-efficients by solving an entropy minimization objective. 
\wemoe~\citep{tang2024merging} merges all modules except for task-specific MLPs, which are retained as weight-ensembled \ac{MoE} with learned routers. 
\twinm~\citep{lu2024twinmerging} leverages \ac{MoE} on difference vectors \ie the difference between the fine-tuned models and the merged model.
While the availability of validation data enhances accuracy, such data might be difficult to obtain in practice.

\textbf{Data-free merging} directly merges model parameters without any data.
\ties~\citep{NEURIPS2023_1644c9af} resolves parameter interference by trimming redundant parameters and resolving sign conflicts. 
\pcbm~\citep{du2024parameter} considers both intra- and inter-parameter competition balancing. 
\dare\citep{yu2024language} reduces parameter interference by randomly dropping parameters and proportionally rescaling remaining ones.
\emr~\citep{huang2024emrmerging} introduces the paradigm of maintaining light-weight task specific masks in addition to the merged model to enhance performance. 
\consensusm~\citep{wang2024localizing} also relies on task specific masks, but creates them differently compared to \emr.
Both approaches significantly improve accuracy over previous methods, albeit at the cost of test-time reconstruction overhead~\citep{gargiulo2025task}. 
\tsvm~\citep{gargiulo2025task} merges SVD-decomposed task singular vectors, reducing interference by retaining only prominent singular directions and orthogonalizing them across tasks.

Recent work by \citet{Zhang_Liu_Ding_Ou_Yu_Zhuang_2025} explores merging into sizes $> 1\times$. 
Their method, \channelm, relies on layer-wise K-Means clustering followed by merging within each cluster using only \tam.
However, this approach is restrictive as it cannot generate fractional-sized models.
Despite the emergence of advanced methods and attempts at merging into bigger sizes, to the best of our knowledge, no prior work has systematically investigated the accuracy–size trade-off in model merging under a single unified framework. 
For completeness, we provide additional related work and a taxonomy of existing algorithms based on their data-free/data-based nature in Appendix~\ref{appendix:extended_rw}.

\section{\sys}

\subsection{Preliminaries}
\label{sec:preliminaries}
We consider a set of $M$ tasks: $\{T_1, \hdots, T_M\}$, where the fine-tuned model parameters for task $T_i$ are denoted by $\btheta_i$. 
These fine-tuned parameters are typically obtained by adapting a pre-trained model, such as ViT~\citep{dosovitskiy2021an} or T5~\citep{raffel2020exploring} using either full parameter \ac{FT} or parameter-efficient fine-tuning (PEFT) methods~\citep{liu2022few}. 
Thus, it is assumed that all the fine-tuned models have the same size and the model architecture as the pre-trained model, as also considered in prior work \citep{ilharco2023editing, NEURIPS2023_1644c9af}.
To analyze the changes introduced by fine-tuning, we use the concept of task vectors $\tv_i$ introduced by \citet{ilharco2023editing}, where $\tv_i = \btheta_i - \btheta_\textrm{pre}$, with $\btheta_\textrm{pre}$ being the pre-trained weights. These task vectors capture the specific modifications needed for each task and provide a compact representation for merging.

Standard model merging approaches involve combining the task-vectors $\{\tv_1, \hdots, \tv_M\}$ into a unified task vector $\tv_\textrm{uni} = \mathcal{F}(\{\tv_1, \hdots, \tv_M\})$ and then adding the unified task vector to the pre-trained weights to get the final merged model, $\btheta_\textrm{uni} = \btheta_\textrm{pre} + \tv_\textrm{uni}$. Here $\mathcal{F}$ denotes the merging algorithm used to obtain the unified task vector's weights. For example, the unified task vector $\tv_\textrm{uni}$ can be computed via simple averaging $\tv_\textrm{uni} = \frac{1}{M} \sum_{i=1}^M \tv_i$ or via \tam~\citep{ilharco2023editing} that uses a coefficient $\lambda$ to weigh the contribution\footnote{We add a scaling factor of $1/M$ to the standard definition $\tv_\textrm{uni} = \lambda \cdot \sum_{i = 1}^M \tv_i$ given in \citep{ilharco2023editing} to better suit its usage in \sys where $M$ can vary across blocks.} of the unified task vector $\tv_\textrm{uni} = \lambda \cdot \frac{1}{M} \sum_{i = 1}^M \tv_i$ in the final merged model. %
It is shown that just by tuning $\lambda$, one can outperform weight averaging~\citep{ilharco2023editing}.

\textbf{Motivation.} Merging into one model $\btheta_\textrm{uni}$ may cause accuracy deterioration due to parameter interference between different fine-tuned models \citep{Zhang_Liu_Ding_Ou_Yu_Zhuang_2025,NEURIPS2023_1644c9af}. 
This behavior becomes prominent as more and more fine-tuned models are merged, as discussed in \Cref{sec:introduction}. On the other hand, retaining all fine-tuned models preserves full fine-tuning accuracy but results in a net size $M \times$ that of one fine-tuned model, which is impractical due to the high memory requirements.
In this work, we investigate the problem of generating models of any desired size in the range \([1,M]\), including models with fractional size such as $2.25\times$ model units. 

\subsection{Proposed approach}

To enable a more granular fusion, we consider the model to be composed of $B$ sequential blocks, for instance transformer blocks in a ViT model or even layers within each transformer block such as attention or MLP layers could be considered as unique blocks.
Assuming $B$ total blocks, we consider the task vectors for each block as $\{\tv^b_k\}_{b=1}^B$ corresponding to the original task vector $\tv_k$ for a task $k$.
Our proposed framework, \sys, takes a greedy approach to efficiently merge task vectors from multiple tasks at the granularity of blocks, aiming to reduce the deployed model size while maintaining utility. 
The pseudo-code for \sys is presented in \Cref{alg:block_merge_algorithm}. 

\textbf{Initialization (Lines 1–6).} The merging proceeds bottom-up. Initially, no merging has occurred, and we retain \( \tv_k^b \) for all tasks \( k \in [M] \) and all blocks \( b \in [B] \) (see \Cref{fig:intro_fig}(a)).
For each block \( b \), we initialize a set of tuples: $\mathcal{G}^b = \left\{ \left( \{k\}, \tv_k^b \right) \mid k \in [M] \right\}$.
Each tuple in \( \mathcal{G}^b \) consists of: \emph{(i)} a task set \( \{k\} \) (tracking which tasks are represented) and \emph{(ii)} the corresponding block task vector \( \tv_k^b \).
For example, in \Cref{fig:intro_fig}(a) for the first block, we would have \( \mathcal{G}^1 = \left\{ \left(\{a\}, \tv_a^1\right), \ldots, \left(\{e\}, \tv_e^1\right) \right\} \).  
When the merging terminates, the resulting $\mathcal{G}^1$ for \Cref{fig:intro_fig}(a) would be \( \mathcal{G}^1 = \left\{ \left(\{a,\ldots,e\}, \hat{\tv}_\textrm{uni}^1\right) \right\} \), where \( \hat{\tv}_\textrm{uni}^1 \) is the merged task vector for the first block for all tasks.
The initial size \(S\) is calculated as the cumulative size of all block parameters across \(M\) tasks. 

\begin{center}

\begin{minipage}[c]{0.94\linewidth}
\centering
\begin{algorithm}[H]
\DontPrintSemicolon
\caption{The \sys framework}
\label{alg:block_merge_algorithm}
\KwIn{Task vectors $\{\tv_k^b\}$ for all $k \in [M], b \in [B]$; merging algorithm $\mathcal{F}$; target size $S_{\text{target}}$}
\KwOut{Merged task vectors with reduced size}

$S \gets 0$ \Comment*[r]{Initialize deployed size}

\For{$b = 1$ \KwTo $B$}{
  $\mathcal{G}^b \gets \emptyset$\;
  \For{$k = 1$ \KwTo $M$}{
    $\mathcal{G}^b \gets \mathcal{G}^b \cup \left(\{k\}, \tv_k^b\right)$\;
    $S \gets S + \text{size}(\tv_k^b)$\;
  }
}

\While{$S > S_{\text{target}}$ \textbf{or} not all blocks merged}{
  Find block $b^*$ and pair $(g_{i^*}, g_{j^*}) \in \mathcal{G}^{b^*}$ with the highest similarity:\;
  \[
    (b^*, g_{i^*}, g_{j^*}) = \argmax_{b \in [B],\, g_i, g_j \in \mathcal{G}^b} \textsc{similarity}(g_i, g_j)
  \]

  $\mathcal{T}_{i^*}^{b^*}, \mathcal{T}_{j^*}^{b^*} \gets g_{i^*}(0),\ g_{j^*}(0)$ \Comment*[r]{Get task subsets}
  $\mathcal{T}_{\text{uni}}^{b^*} \gets \mathcal{T}_{i^*}^{b^*} \cup \mathcal{T}_{j^*}^{b^*}$ \Comment*[r]{Merge task subsets}
  $\tv_{\text{uni}}^{b^*} \gets \mathcal{F}(\{ \tv_k^{b^*} \mid k \in \mathcal{T}_{\text{uni}}^{b^*} \})$ \Comment*[r]{Merge task vectors}
  
  $\mathcal{G}^{b^*} \gets \mathcal{G}^{b^*} \cup \left(\mathcal{T}_{\text{uni}}^{b^*}, \tv_{\text{uni}}^{b^*}\right) \setminus \{g_{i^*}, g_{j^*}\}$ \Comment*[r]{Update the block}
  
  $S \gets S - \text{size}(\tvmtl^{b^*})$ \Comment*[r]{Update current size}
}
\end{algorithm}
\end{minipage}
\end{center}

\textbf{Iteration (lines 7-14).} In each iteration, the algorithm identifies a block $b^*$ and pair of tuples $(g_{i^*}, g_{j^*}) \in \mathcal{G}^{b^*}$, which have the highest similarity (as defined below). Then they are merged as follows. Let $\mathcal{T}_{i^*}^{b^*}$ and $\mathcal{T}_{j^*}^{b^*}$ be the subset of tasks associated with $g_{i^*}$ and $g_{j^*}$ respectively, \ie the first elements of $g_{i^*}$ and $g_{j^*}$ respectively. First, $\mathcal{T}_{i^*}^{b^*}$ and $\mathcal{T}_{j^*}^{b^*}$ are merged via a union operation: $\mathcal{T}_{\text{uni}}^{b^*} = \mathcal{T}_{i^*}^{b^*} \cup \mathcal{T}_{j^*}^{b^*}$. Next, the merged task vector corresponding to block $b^*$ and set $\mathcal{T}_{\text{uni}}^{b^*}$ is created as follows: $\tv_{\text{uni}}^{b^*} = \mathcal{F}(\{ \tv_k^{b^*} \mid k \in \mathcal{T}_{\text{uni}}^{b^*}\})$. 
Here $\mathcal{F}$ can be \textit{any} data-free merging algorithm.
The tuple set $\mathcal{G}^{b^*}$ is then updated by removing the tuples $g_{i^*}, g_{j^*}$ and adding the new merged tuple $(\mathcal{T}_{\text{uni}}^{b^*}, \tv_{\text{uni}}^{b^*})$. Each merge reduces the model size by the size of the task vector corresponding to block \(b^*\), and the process continues until the current size \(S\) meets the desired size \(S_{\text{target}}\) or no further merges are possible. 

\textbf{Similarity function.} 
We measure the similarity between two groups \(g_i, g_j\) in any block $b$ using the lowest cosine similarity between any pair of original task vectors corresponding to the tasks in the sets $\mathcal{T}_{i}^{b}$ and $\mathcal{T}_{j}^{b}$:
\begin{equation}
\label{eqn:similarity}
\textsc{similarity}(g_i, g_j) = \min_{k_1 \in \mathcal{T}_i^{b}, \text{ } k_2 \in \mathcal{T}_j^{b}} \text{cosine\_sim}(\tv^b_{k_1}, \tv^b_{k_2}).    
\end{equation}
Our choice of the $\min$ similarity derives from our ablations comparing different strategies—$\max, \min$, and average—as well as computing similarity between merged group task vectors directly.
Among these, $\min$ yields the best performance.
Thus at each iteration, we merge the pair of groups with the highest of these minimum similarities (line 9, \Cref{alg:block_merge_algorithm}).
While the cosine similarity between full task vectors can be relatively low~\citep{ilharco2023editing}, the block-level similarities tend to be higher and effective for merging. \channelm~\citep{Zhang_Liu_Ding_Ou_Yu_Zhuang_2025} also employs cosine similarity. 

\textbf{Enhancing efficiency.} The pairwise similarities can be precomputed once for all pairs and accessed in constant time during the merging process. 
Furthermore, we implement $\mathcal{G}^b$ using the Disjoint Set Union (DSU)~\citep{10.5555/1614191} data structure to efficiently track and unify task sets for each block.
Our design enables \sys to perform very efficient merging even under many tasks (see \Cref{tab:merging_time}).

\textbf{Complexity Analysis.} \Cref{alg:block_merge_algorithm} identifies most similar block pairs in each iteration. We presented the algorithm in this form for conceptual clarity. 
However, in practice, we can generate the global merging order for all blocks first and then apply merges.  
To analyze the time complexity of \sys, we consider three distinct stages of this process: similarity pre-computation, generating merging order, and actual parameter merging. 

\begin{itemize}[leftmargin=10pt] 
    \item \textbf{Similarity pre-computation:} We compute the pairwise cosine similarities between $M$ tasks in each block, across all $B$ blocks.  Let the maximum size of any block task vector be $d_\text{max}$, then the similarity computes takes $\mathcal{O}(d_\text{max})$ per pair. With $\binom{M}{2}$ pairs per block, this step is $\mathcal{O}(BM^2 d_\text{max})$. 
    
    \item \textbf{Generating merging order:} Our greedy merging using \Cref{eqn:similarity} is an instance of a specific form of clustering, called single-linkage clustering. We thus use the SLINK algorithm~\citep{SLINK} which takes as input the similarity matrix and generates a sorted list of merge orders for each block in $\mathcal{O}(M^2)$. For $B$ blocks, this takes $\mathcal{O}(BM^2)$. We then need to combine these per-block sorted lists, each of size $M-1$, into a single global sorted list.  Using a min-heap, this takes $\mathcal{O}(BM\log B)$~\citep{knuth1997art}. In total, this step takes $\mathcal{O}(BM^2 + BM\log B)$ and results in a global merge ordering across all blocks. 
    Once the global merge ordering is obtained, we replay merges until the target size is met, simultaneously tracking task clusters via a DSU data structure, one per block. 
    This gives the final groups $\{\mathcal{G}^b\}_{b=1}^B$ to which parameter merging is then applied. 
    The per-block DSU overhead is $\mathcal{O}(M\alpha(M))$, where $\alpha(\cdot)$ is the Inverse Ackermann Function~\citep{10.5555/1614191}. 
    For all practical purposes, $\alpha(M) < 5$, resulting in $\approx \mathcal{O}(BM)$ for $B$ blocks.
    
    \item \textbf{Applying parameter merging:} We now merge parameters according to the final groupings in $\mathcal{G}^b$. For linear algorithms like TA, merging a group $g$ of size $|g|$ with task vectors of size $d$ takes $\mathcal{O}(|g|d)$. Summing over all groups within a block takes $\sum_{g \in \mathcal{G}^b} \mathcal{O} (|g|\cdot d) = \mathcal{O} ((\sum_{g \in \mathcal{G}^b}|g|)d) = \mathcal{O} (Md)$. Repeating this for $B$ blocks and upper bounding $d$ with $d_\text{max}$ results in $\mathcal{O}(BMd_\text{max})$.
\end{itemize}

The total complexity is dominated by the similarity pre-computation (as $d_\text{max}$ is typically larger than $B$), resulting in a final complexity of $\mathcal{O}(BM^2 d_\text{max})$. Note however that $d_\text{max}$ is much smaller than the total model dimension, as it only corresponds to the maximum size of any block of the model. 

\begin{table}[t!]
     \centering 
      \caption{Summary of existing data-free merging methods. Column $\mathcal{F}(\{ \tv_k^{b} \mid k \in \mathcal{T}_{\text{uni}}^{b}\})$ denotes the result of merging. \Cref{fig:flexmerge_plus_algs} (\Cref{appendix:additional_details}) provides an illustrative diagram.}
     \label{tab:merging_algorithms_details}
     \resizebox{\textwidth}{!}{%
     \begin{tabular}{c@{\hspace{0.2cm}} c c c}
        \toprule
         \textbf{Algorithm} & \textbf{$\mathcal{F}(\{ \tv_k^{b} \mid k \in \mathcal{T}_{\text{uni}}^{b}\})$} & \textbf{Final Model} & \textbf{What is stored?} \\
         \midrule
          \makecell{\tam~\citep{ilharco2023editing}, \\
          \textsc{TIES}~\citep{NEURIPS2023_1644c9af}, \\ Avg.~\citep{ilharco2023editing}, \\
          \textsc{PCB}~\citep{du2024parameter}, \\
          \textsc{TSV-M}~\citep{gargiulo2025task}} & $\tv_\textrm{uni}^{b}$ & $\btheta_\textrm{uni}^{b} = \btheta_\textrm{pre}^{b} + \tvmtl^{b}$ & $\btheta_\textrm{uni}^{b}$ \\
          \midrule
          
          \consensusm~\citep{wang2024localizing} 
          & 
          $\tv_\textrm{uni}^{b}, \{\mask_k^{b} \mid k \in \mathcal{T}_{\text{uni}}\}$ 
          & \makecell{$\bthetah_k^{b} = \btheta_\textrm{pre}^{b} + \tvmtl^{b} \circ \mask_{k}^{b}$ \\ (reconstructed per-task $k$)} 
          & 
          
          \makecell{$\btheta_\textrm{pre}^{b}, \tvmtl^{b},$ \\ $ \{\mask_{k}^{b} \mid k \in \mathcal{T}_{\text{uni}}\}$}
          
          \\
          \midrule
          
          \emr~\citep{huang2024emrmerging} & 
          $\tv_\textrm{uni}^{b}, \{\mask_k^{b}, \gamma_k^{b} \mid k \in \mathcal{T}_{\text{uni}}^{b}\}$
          & \makecell{$\bthetah_k^{b} = \btheta_\textrm{pre}^{b} + \gamma_k^{b} \cdot \tvmtl^{b} \circ \mask_{k}^{b}$ \\ (reconstructed per-task $k$)} & 
          \makecell{$\btheta_\textrm{pre}^{b}, \tvmtl^{b},$ \\ $ \{\mask_k^{b}, \gamma_k^{b} \mid k \in \mathcal{T}_{\text{uni}}\}$}
          \\
         \bottomrule 
     \end{tabular}
     }
 \end{table}

\subsection{Existing merging methods in combination with \sys}
\label{subsec:sec3_merging_algs}

\sys provides the flexibility to choose any data-free merging algorithm $\mathcal{F}$ from a diverse set of existing approaches.
Unlike traditional methods that operate at the level of full task vectors, \sys applies merging algorithms at the block level, fusing block task vectors. 
We detail the exact block-level merging procedure for different algorithms next.
In standard approaches like \tam, \tsvm, and \pcbm, task vectors are merged into a single unified task vector. 
When applied at the block-level, the merging outcome for any block $b$ can be denoted as: \mbox{$\tvmtl^b \leftarrow \mathcal{F}(\{ \tv_k^{b} \mid k \in \mathcal{T}_{\text{uni}}^{b}\})$} where $\mathcal{F}$ is the specific merging algorithm and $\mathcal{T}_{\text{uni}}^{b}$ is the subset of tasks for which the merging occurs.
The final block parameters are then computed as $\btheta_\textrm{uni}^{b} = \btheta_\textrm{pre}^{b} + \tvmtl^{b}$.
Approaches such as \consensusm generate task-specific masks in addition to the unified vector: $\tvmtl^b, \{\mask_k^{b} \mid k \in \mathcal{T}_{\text{uni}}\} \leftarrow \mathcal{F}(\{ \tv_k^{b} \mid k \in \mathcal{T}_{\text{uni}}^{b}\})$.
Then during inference, the task-specific weights for task $k$ are reconstructed as $\bthetah_k^{b} = \btheta_\textrm{pre}^{b} + \tvmtl^{b} \circ \mask_{k}^{b}$.
\consensusm thus stores $\btheta_\textrm{pre}^{b}$, $\tvmtl^{b}$, and the binary masks $\{\mask_{k}^{b} \mid k \in \mathcal{T}_{\text{uni}}\}$ and defers per-task reconstruction to the inference time.
This leads to a storage cost exceeding $2\times$ that of standard methods, which only store $\btheta_\textrm{uni}^b$.
\emr further generates task-specific scalars $\{\gamma_{k}^{b} \mid k \in \mathcal{T}_{\text{uni}}\}$ in addition to the masks, however the storage cost of these scalars is negligible.
\Cref{tab:merging_algorithms_details} summarizes the merging outcomes for different algorithms, applied at block-level within \sys. 
\Cref{fig:flexmerge_plus_algs} (\Cref{appendix:additional_details}) provides an illustrative diagram. 

\section{Experiments}
\label{sec:experiments}

We split our evaluation as follows: \emph{(i)} Merging on vision, \ac{PEFT} and \ac{FFT} benchmarks in \Cref{subsec:merging_results}; (ii) \sys vs \channelm in \Cref{subsec:compare_to_channelmerge}; and \emph{(iii)} ablation and efficiency analysis in \Cref{subsec:analysis}. Lastly, multi-modal and OOD results are in Appendices~\ref{subsec:appendix_multimodal_exps} and ~\ref{subsec:appendix_OOD}.

\textbf{Merging algorithms.} We investigate the accuracy-size trade-off for several data-free merging algorithms including Vanilla Averaging, \tam~\citep{ilharco2023editing}, \ties~\citep{NEURIPS2023_1644c9af}, \pcbm~\citep{du2024parameter}, \tsvm~\citep{gargiulo2025task}, \wudim~\citep{cheng2025whoever}, \consensusm~\citep{wang2024localizing} and \emr~\citep{huang2024emrmerging} on extensive vision and NLP benchmarks. 
As noted earlier, the focus of our work is data-free model merging.
Hence, existing data-based algorithms such as \surgery~\citep{pmlr-v235-yang24t}, \adam~\citep{yang2024adamerging}, \twinm~\citep{lu2024twinmerging}, \etc are not directly comparable in our setting.

\begin{figure}[t!]
    \centering
    \includegraphics[]{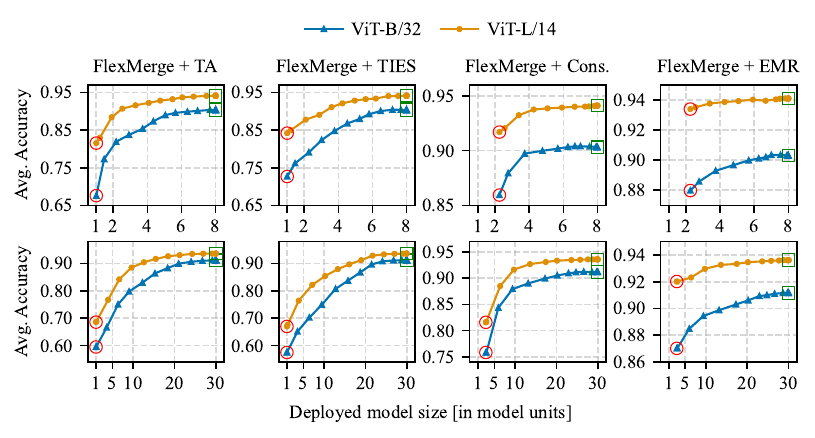}
    \caption{Merging $8$ (top) and $30$ (bottom) tasks. The accuracy-size trade-off shows rapid initial gains, followed by gradual improvement, reaching near fine-tuning accuracy well before the maximum size.
    }
    \label{fig:vision}
\end{figure}

\textbf{Hyperparameters.} For \tam, we set $\lambda = 1.5$. 
For \ties, we use a sparsity ratio of $0.1$ and employ the recommended value of $\lambda=1$. 
For \consensusm, we set the hyperparameter responsible for controlling the amount of information extracted by masks to $0.6$ for all tasks and use \ties as the algorithm to generate unified task vectors.
For \sys, we set the block granularity at the level of individual components within the transformer layer, \ie the attention, MLP, and layer normalization modules are treated as separate blocks during the merging process.

\subsection{Merging Results}
\label{subsec:merging_results}

\textbf{Merging \(8\) and \(30\) vision models.}
For the image classification tasks, we follow the setup from existing work~\citep{huang2024emrmerging,NEURIPS2023_1644c9af}. Specifically, we use two versions of the CLIP model~\citep{radford2021learning}, incorporating ViT-B/32 and ViT-L/14 as visual encoders~\citep{dosovitskiy2021an}. 
We evaluate on the standard $8$ task benchmark~\citep{ilharco2023editing} as well as an extended $30$ task benchmark (detailed in Appendix~\ref{subsec:appendix_vision_benchmark_details}). 
\Cref{fig:vision} plots average accuracy vs. deployed model size (in multiples of a single fine-tuned model).
For \sys + \tam, the accuracy increases fairly rapidly as the model size grows beyond $1\times$.
The gains are significant (top row), where the accuracy reaches $> 80\%$ at size $2\times$ from only $67.5\%$ at size $1\times$ for the \vitb model in the $8$ task setup. 
Similar gains are also observed for $30$ tasks (bottom row).

Masking-based approaches, \consensusm and \emr, begin with substantially higher accuracy than \tam and \ties, but their smallest size exceeds $1\times$ due to the need to store pre-trained weights and binary masks (\Cref{subsec:sec3_merging_algs}).
On $8$ tasks, \consensusm was shown to match fine-tuned accuracy at small sizes, but only when its extraction parameter is separately tuned per task~\citep{wang2024localizing}. 
\sys + \consensusm also shows strong gains, improving from $76\%$ at $\approx 3\times$ to $84.5\%$ at $\approx 6\times$ for \vitb in $30$ tasks.
\emr maintains high accuracy even at the smallest size.
Yet, it exhibits an accuracy gap w.r.t the fine-tuned models, which can be effectively reduced by increasing the deployed model size. 
Larger \vitl models achieve higher accuracy across all methods, but the accuracy-size trade-off remains similar: rapid initial gains followed by gradual improvements.
Most algorithms approach the fine-tuning accuracy (denoted by $\square$) well before maximum size, around $6\times$ for $8$ tasks and $23.5\times$ for $30$ tasks.
Thus, in cases requiring storage of all fine-tuned models, \sys can reduce size by about $25\%$ with little accuracy loss.

\begin{figure}[t]
    \centering
    \includegraphics[]{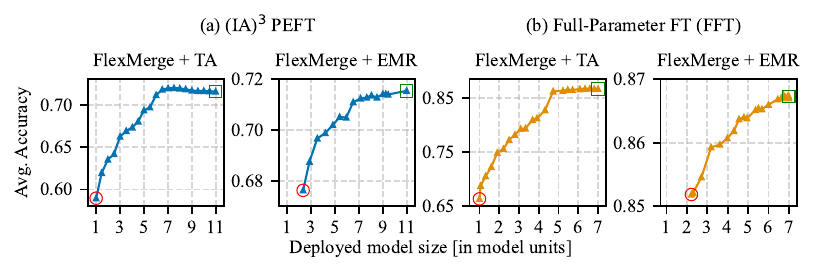}
    \caption{\sys + \tam gains $7.2\%$ for (IA)$^3$ going from $1\times$ to $3\times$ and more than $9\%$ for FFT when just doubling the size from $1\times$ to $2\times$. \textsc{EMR} begins with higher accuracy, yet, substantially benefits from increased size.}
    \label{fig:nlp_all}
\end{figure}

\textbf{Merging $11$ \ac{PEFT} models.} 
We adopt the experimental setup from prior work ~\citep{huang2024emrmerging,NEURIPS2023_1644c9af}. 
Specifically, we employ the (IA)\(^3\)~\citep{liu2022few} PEFT method on the T0-3B~\citep{sanh2022multitask} base model using $11$ diverse datasets sourced from \citep{NEURIPS2023_1644c9af} (detailed in \Cref{subsec:appendix_peft_details}). 
\Cref{fig:nlp_all}(a) demonstrates the benefits of deploying larger model sizes, where in this case the model size is measured with respect to the (IA)$^3$ modules.
\sys + \tam achieves notable gains, increasing accuracy from $59\%$ at size $1\times$ to $66.2\%$ at $3\times$, a $7.2\%$ improvement. Similarly, \sys + \emr surpasses $70\%$ accuracy at $5\times$, starting from $67.6\%$ at the lowest size of $2.34\times$.
We observe similar trends for other algorithms, included in Appendix~\ref{subsec:appendix_peft_exps}.

\textbf{Merging $7$ FFT models.} 
For this experiment, we closely follow the setup from prior work~\citep{du2024parameter,NEURIPS2023_1644c9af}. We use \tfivebase and \tfivelarge as base models, applying full-parameter fine-tuning on $7$ datasets sourced from \citep{NEURIPS2023_1644c9af} (detailed in  Appendix~\ref{subsec:appendix_full_parameter_ft_details}). \Cref{fig:nlp_all}(b) illustrates the trade-off between model size and accuracy for the \tfivelarge model. Here, one unit of model size corresponds to the full size of a single model. 
\sys + \tam gains more than $9\%$ to reach an accuracy of $75\%$ when just doubling the size from $1\times$ to $2\times$.
Similarly, \sys + \emr surpasses $86\%$ at size $4\times$, starting from $85.2\%$ at its lowest size of $2.2\times$. 
Consistent with our observations on vision tasks, \sys + \tam reaches very close to the fine-tuning accuracy around size $5\times$, much in advance of full size $7\times$.
Thus, scaling the model size benefits both ends of the spectrum.
Results for other combinations are included in Appendix~\ref{subsec:appendix_fft_exps}.

\begin{figure}[t!]
    \centering
    \includegraphics[]{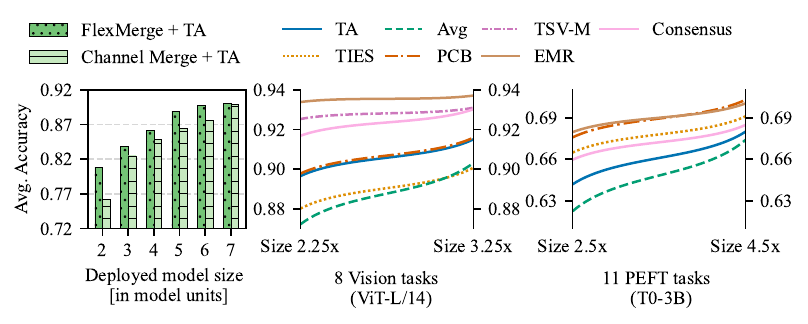}
    \caption{
    (Left) \sys + \tam outperforms \channelm + \tam across all sizes. 
    (Center, Right) Algorithm rankings shift even at modestly larger sizes, with simpler methods rivaling advanced ones. 
    We show sizes just over \consensusm and \emr's lowest size for a wholistic comparison.
    }
    \label{fig:baseline_plus_switching}
\end{figure}

\textbf{Cross-algorithm analysis.} Thus far, we evaluated the accuracy-size trade-off per algorithm. 
We now compare algorithms at same size, yielding two interesting findings: \emph{(i)} the performance gap between different algorithms significantly narrows at slightly larger sizes; and \emph{(ii)} the algorithms rankings also alter in many cases, with simpler algorithms rivaling or surpassing advanced ones.
This behavior aligns with the phenomenon we observed earlier: as scale increases, parameter interference between tasks reduces substantially, leading to greater natural parameter disentanglement and higher accuracy. Consequently, the explicit interference-reduction mechanisms built into advanced approaches such as TRIM in TIES or competition balancing in PCB offer marginal added benefit, because much of the interference is already mitigated simply by the increase in scale. Thus simple algorithms begin rivaling their more sophisticated counterparts at larger sizes.
In \Cref{fig:baseline_plus_switching} on vision tasks, vanilla averaging exceeds \ties at size $3.25\times$ while \tam overlaps with \textsc{PCB}.
While \emr and \consensusm stay atop on vision, they are surpassed by \textsc{PCB} on \ac{PEFT} at size $4.5\times$. 
Crucially, all algorithms remain within $3-4\%$ on both benchmarks at increased sizes despite originating with a much larger gap at size $1\times$ (see \Cref{fig:intro_fig}(b)). 
\textit{Our findings provide encouraging evidence to develop and compare algorithms at sizes $>1\times$ rather than only at $1\times$.}

\subsection{\sys vs \channelm}
\label{subsec:compare_to_channelmerge}
\channelm ~\citep{Zhang_Liu_Ding_Ou_Yu_Zhuang_2025} uses K-Means clustering per layer, following a fixed same value of $K$ for every layer.
Each choice of $K \in \{2, 3, \ldots, M-1\}$ results in a merged model of the corresponding size. 
\Cref{fig:baseline_plus_switching} charts the average accuracy with \tam and \vitb for a set of integer model sizes, excluding the extremes $1\times$ and $8\times$ where both approaches have the identical accuracy.
Recall that \channelm does not support fractional sizes.
\sys achieves higher accuracy than \channelm in all cases, thanks to its greedy pairwise merging approach which allows flexible number of groups per layer instead of restrictive clustering.  
Results with \ties and visualization of clusters is included in Appendix~\ref{subsec:appendix_flex_vs_channel}.

\subsection{Analysis}
\label{subsec:analysis}

\begin{figure}[t!]
    \centering
    \includegraphics[]{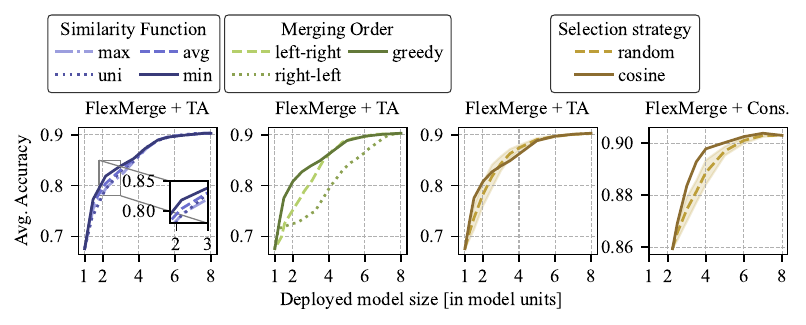}
    \caption{
    Ablation results for \sys reveal that the min similarity strategy and greedy merging perform the best, while cosine similarity generally outperforms random selection.}
    \label{fig:merging_ablations}
\end{figure}

\textbf{Ablations on the merging procedure.}
We ablate on the similarity functions (min, max, average, comparing unified vectors) for \Cref{eqn:similarity} and merging orders (left-to-right, right-to-left, greedy) in \sys using the \vitb model on $8$ tasks.
We also investigate random block selection over cosine similarity.
\Cref{fig:merging_ablations} shows that the min strategy performs the best, though other strategies are also competitive.
For merging order, right to left performs the worst as expected since the final layers in neural networks tend to be more specialized and merging them first hurts accuracy.
While left to right seems ideal, it can be too strict and therefore greedy emerges as the best.
We further analyze the merging order of greedy in Appendix~\ref{subsec:merging_order_analysis}.
Random selection is competitive, but generally underperforms when compared across algorithm. 
Based on these findings, we set \sys to use greedy with cosine similarity (min strategy) by default.
For more ablations, see Appendix~\ref{subsec:appendix_ablations}.

\textbf{Merging and inference efficiency.}
\Cref{tab:merging_time} shows that \sys achieves highly efficient data-free merging, generating all deployed sizes in about \SI{20}{\sec} for up to 30 tasks.
For inference with \sys, each request follows a unique forward path through the merged model using task-specific blocks (\Cref{fig:intro_fig}(a)). For a model of size $1\times$, all tasks share a single path, but the classification heads are always applied separately.
We load the tensors of merged model (size $> 1\times$) into the GPU memory once and create $M$ task-specific model views that reference these shared tensors to process task batches in parallel. 
Standard merging, by contrast, processes all tasks in a single batch before splitting for task-specific heads.
We simulate the worst case arrival, where inference batches corresponding to all tasks arrive at once.
We consider $50$ consecutive batches of size $256$ (totaling $12800$ samples).
Each batch contains $32$ samples per task across $8$ tasks.
\Cref{tab:inference_measurements} shows that \sys maintains inference speed comparable to standard merging for both \vitb and \vitl, demonstrating that larger models can enhance accuracy without slowing inference.

\begin{table}[h!]
    \centering
    \begin{minipage}[t]{0.38\textwidth}
        \centering
        \caption{\sys's merging time.}
        \label{tab:merging_time}
        
        \scriptsize 
        \setlength{\tabcolsep}{2.5pt} 
        
        \begin{tabular}{l c c} 
            \toprule
            \textbf{Method} & \textbf{\vitb (s)} & \textbf{\vitl (s)} \\
            \midrule
            
            \multicolumn{3}{c}{\textit{8 Tasks}} \\ 
            \cmidrule(lr){1-3}
            \tam (size $1\times$) & $\approx 0.8$ & $\approx 2.6$ \\
            \sys (all sizes)      & $\approx 2.3$ & $\approx 3.4$ \\
            
            \midrule
            
            \multicolumn{3}{c}{\textit{30 Tasks}} \\ 
            \cmidrule(lr){1-3}
            \tam (size $1\times$) & $\approx 1.9$ & $\approx 6.1$ \\
            \sys (all sizes)      & $\approx 20$  & $\approx 31$   \\
            \bottomrule
        \end{tabular}
    \end{minipage}
    \hfill
    \begin{minipage}[t]{0.6\textwidth}
        \centering
        \caption{Comparing inference time of \sys against standard model merging. The overheads are negligible.}
        \label{tab:inference_measurements}
        \resizebox{\textwidth}{!}{
        \begin{tabular}{c c c c}
        \toprule
            \textbf{Model} & \textbf{Algorithm} & \textbf{Size} & \textbf{Inference Cost (/$12800$ items)} \\
            \cmidrule(lr){1-1}\cmidrule(lr){2-4}
            \multirow{2}{*}{ViT-B-32} & Standard Merging & $1\times$ & $12.30 \pm 0.21$ ms\\
            & \sys & $> 1\times$ & $12.21 \pm 0.41$ ms \\
            \midrule
            \multirow{2}{*}{ViT-L-14} & Standard Merging & $1\times$ & $118.70 \pm 1.78$ ms\\
            & \sys & $> 1\times$ & $120.53 \pm 0.32$ ms\\
        \bottomrule
        \end{tabular}
        }
    \end{minipage}
\end{table}

\section{Discussion and Conclusion}
\label{sec:discussion_conclusion}

\textbf{Benefits.} Different merging algorithms have different advantages: \textsc{EMR} and \consensusm achieve high accuracy but require task-specific reconstruction during inference, incurring overheads. \sys can also mitigate this overhead as larger deployed models need fewer blocks to be reconstructed (see Appendix~\ref{subsec:appendix_reconstruction_latency}).
In contrast, \textsc{TIES} and \tam avoid reconstruction but have lower accuracy. 
\sys provides flexibility, letting practitioners choose algorithms and balance accuracy, reconstruction overhead, and model size for various deployment scenarios.

\textbf{Limitations.} Most works, including \sys, are limited to merging models with the same architecture as merging heterogeneous models remains challenging~\citep{NEURIPS2020_fb269786,imfeld2024transformer}.
Secondly, the theoretical insights for effective model merging are limited~\citep{ortiz-jimenez2023task}.
For \sys, how to obtain the optimal merged model for any given size remains unclear. 
Although extensive ablations help guide (\Cref{subsec:analysis}), further investigation is needed to understand the bounds of the accuracy-size trade-off.

We introduced \sys, a flexible, data-free model merging framework that extends beyond traditional single-model fusion and offers precise control over fused model size. 
Extensive experiments show that the accuracy-size trade-off exhibits favorable properties for several algorithms, benefiting from rapid accuracy gains with modest size increments.  
Future work may explore specialized algorithms for block-level merging.

\subsubsection*{Acknowledgments}
Akash was supported by the EPFL Doc.Mobility fellowship during his research visit to Carnegie Mellon University.
This work was also partially supported by the NSF grants CCF 2045694, CCF 2428569, CNS-2112471, CPS-2111751, ONR grant N00014-23-1-2149, AI2C Seed grant, and by the Swiss National Science Foundation, under the project ``FRIDAY: Frugal, Privacy-Aware and Practical Decentralized Learning'', SNSF proposal No. 10.001.796.
The authors are thankful to Martijn de Vos and Arian Raje for their helpful feedback during the preparation of this work.

\bibliography{references}
\bibliographystyle{iclr2026_conference}

\clearpage

\appendix

\section*{Organization of the Appendix}
\begin{enumerate}[label={}]
    \item \ref{appendix:extended_rw} Extended related work
    \item \ref{appendix:additional_details} Additional details
    \begin{enumerate}[label={}]
        \item \ref{subsubsec:how_to_apply_at_blocklevel} How to apply existing merging algorithms at the block-level?
        \item \ref{subsec:appendix_vision_benchmark_details} Vision benchmark
        \item \ref{subsec:appendix_peft_details} \Acf{PEFT} benchmark
        \item \ref{subsec:appendix_full_parameter_ft_details} Full-parameter fine-tuning benchmark
        \item \ref{subsec:appendix_muttimodal_details} Multi-modal benchmark
    \end{enumerate}
    \item \ref{sec:appendix_results} Additional results
    \begin{enumerate}[label={}]
        \item \ref{subsec:appendix_vision_exps} Vision benchmark
        \item \ref{subsec:appendix_peft_exps} \Acf{PEFT} benchmark
        \item \ref{subsec:appendix_fft_exps} Full-parameter fine-tuning benchmark
        \item \ref{subsec:appendix_multimodal_exps} Multi-modal benchmark
        \item \ref{subsec:appendix_flex_vs_channel} \sys vs \channelm
        \item \ref{subsec:appendix_OOD} OOD performance of \sys
        \item \ref{subsec:appendix_scaling_laws} \textit{Scaling laws} for flexible model merging
        \item \ref{subsec:appendix_ablations} Ablations of the merging procedure
        \item \ref{subsec:appendix_reconstruction_latency} Reconstruction latency of masking based approaches
        \item \ref{subsec:merging_order_analysis} Merging order analysis
        \item \ref{subsec:appendix_normalized_accuracy} Main results presented with normalized accuracy
        \item \ref{subsec:appendix_datasetwise_results} Dataset-wise results
    \end{enumerate}
    \item \ref{sec:compute_resources} Compute Resources
    \item \ref{sec:llm_usage} LLM Usage Statement
\end{enumerate}

\section{Extended Related Work}
\label{appendix:extended_rw}

\Cref{tab:related_work} characterizes different merging algorithms on their data-based/data-free nature and merged model size.
Below we discuss additional related work.

\textbf{Multi-Task Learning (MTL).}
The traditional approach to obtaining a model with multi-task capabilities is \ac{MTL} which trains a single model using training data from multiple tasks together~\citep{vandenhende2021multi,sanh2022multitask}.
However, \ac{MTL} suffers from not only (i) the expensive computational cost for training, but also (ii) the limited data availability due to data privacy~\citep{jin2023dataless,yang2024adamerging}.
In comparison, model merging bypasses these challenges by combining the fine-tuned model weights directly, without training data, thus offering a more cost-effective approach to building a multi-task model.
Existing research in \ac{MTL} also focuses on grouping tasks \ie identifying subsets of task that derive positive benefit from training together~\citep{standley2020tasks,fifty2021efficiently}. 
Since training one model for all tasks can lead to suboptimal performance due to task conflicts and competition for model capacity, these methods train separate multi-task models for specific task groups.
This can be seen as conceptually similar to our approach of having different task subsets per merged block to improve performance (see \Cref{fig:intro_fig}(a)).
While our approach performs grouping during merging, these approaches perform grouping during training.

\textbf{Model Merging.} Besides the works discussed in \Cref{sec:related_work}, recent
works also focus on merging models fine-tuned specifically with Low-Rank Adaptation (LoRA)~\citep{stoica2025knots,zhao2025lego,tangparameter}. Other efforts focus on developing approaches for fine-tuning that result in lower interference during downstream merging~\citep{lee2025mitigating,jin2025fine,ortiz-jimenez2023task}. 
\linesm~\citep{wang2025lines} proposes a post-training editing technique to reduces negative interference between parameters by scaling parameter updates based on their layer depth.
\sys can seamlessly incorporate these recent advances.

\begin{table*}[h!]
    \centering
    \caption{Comparison of merging algorithms by data dependency and merged model size.}
    \label{tab:related_work}
	\resizebox{\textwidth}{!}{%
	\begin{tabular}{l c c c}
		\toprule
	 \textbf{Algorithm}	& \textbf{Data Free} & \textbf{Size} & \textbf{Storage beyond the unified model}\\
		\midrule
		Weight Average & \cmark & $1\times$ (Fixed) & --\\
	    \tam~\citep{ilharco2023editing} &  \cmark & $1\times$ (Fixed) & --\\
		\ties~\citep{NEURIPS2023_1644c9af} & \cmark & $1\times$ (Fixed)& --\\
            \pcbm~\citep{du2024parameter} & \cmark & $1\times$(Fixed) & -- \\
		 \consensusm~\citep{wang2024localizing} & \cmark & $> 2\times$ (Fixed) & Stores masks and $\btheta_\textrm{pre}$\\
		\emr~\citep{huang2024emrmerging} & \cmark & $> 2\times$ (Fixed) & Stores masks and $\btheta_\textrm{pre}$\\
        \tsvm~\citep{gargiulo2025task} & \cmark & $1\times$ (Fixed) & -- \\
        \wudim~\citep{cheng2025whoever} & \cmark & $1\times$ (Fixed) & -- \\
        \midrule
		\regmean~\citep{jin2023dataless} & \xmark & $1\times$ (Fixed)& --\\
		\fisher~\citep{matena2022merging} & \xmark & $1\times$ (Fixed) & --\\
		 \adam~\citep{yang2024adamerging} & \xmark & $1\times$ (Fixed)& --\\
		\surgery~\citep{pmlr-v235-yang24t} & \xmark & $ > 1 \times$ (Fixed) & Stores task-specific adapters \\
		\wemoe~\citep{tang2024merging} & \xmark & $\gg 1\times$ (Fixed) & Stores MLP modules for all tasks\\
		\twinm~\citep{lu2024twinmerging} & \xmark & $> 2\times$ (Fixed) & Stores compressed diff. vectors and $\btheta_\textrm{pre}$\\
		\bottomrule
	\end{tabular}
    }
    
\end{table*} %
\section{Additional Details}
\label{appendix:additional_details}

\begin{figure}[h!]
    \centering
    \includegraphics[scale=0.25]{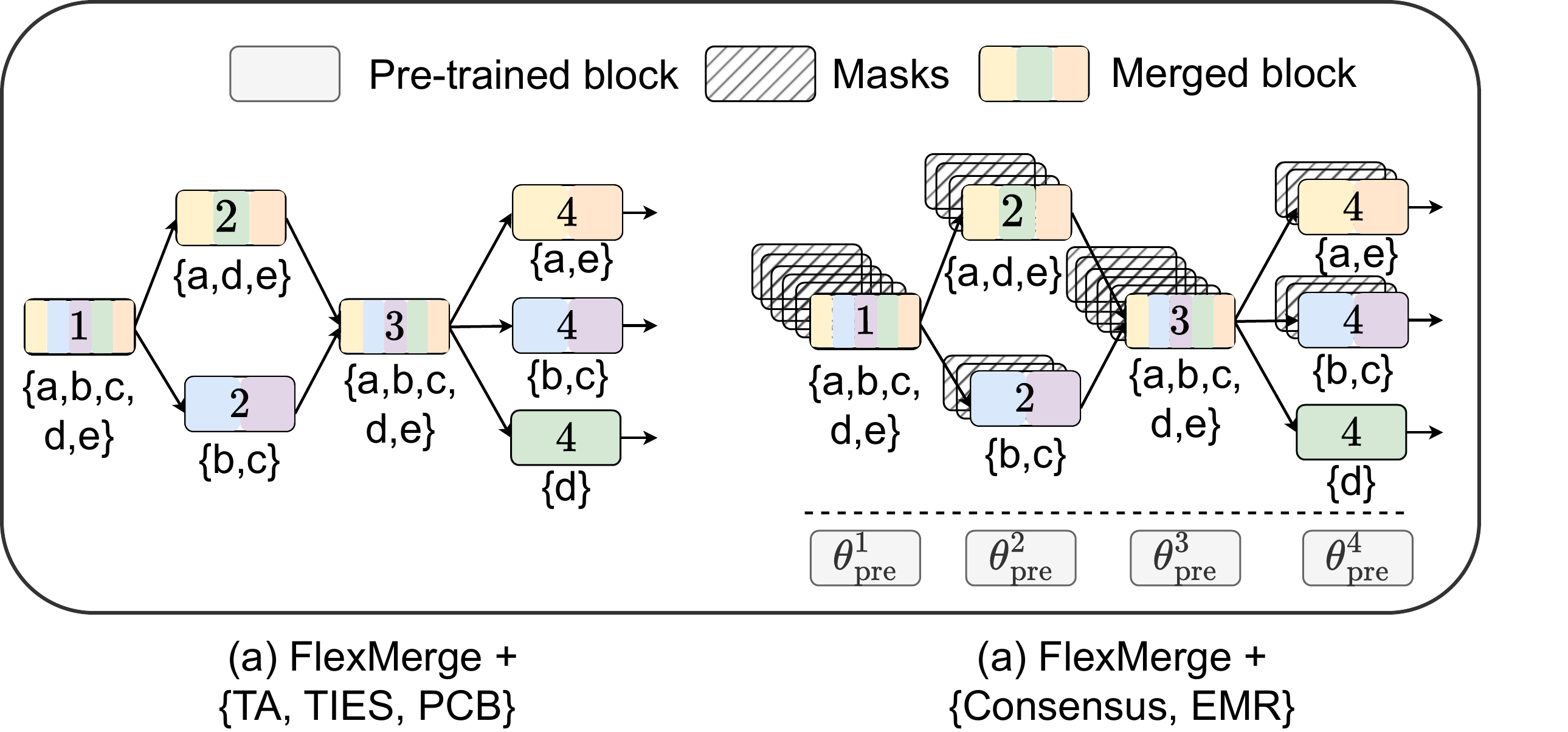}
    \caption{\sys in combination with different merging algorithms. Standard methods such as \tam, \ties, \etc  generate the merged parameters per block. Recent methods such as \consensusm and \emr generate binary masks in addition to the unified parameters. They also store the pre-trained parameters for task-specific reconstruction.}
    \label{fig:flexmerge_plus_algs}
\end{figure}

\subsection{How to apply existing merging algorithms at the block-level?}
\label{subsubsec:how_to_apply_at_blocklevel}

While most merging algorithms can be directly applied at the block-level, we list them below and explain any specific adaptations that help improve performance. \Cref{fig:flexmerge_plus_algs} provides an illustrative diagram for \sys in combination with different merging algorithms. 
\begin{itemize}
    \item \textbf{Averaging}. Applied directly to obtain $\tv_\textrm{uni}^b$. 
    \begin{align*}
        \mathcal{F}(\{ \tv_k^{b} \mid k \in \mathcal{T}_{\text{uni}}^b\}) = \frac{1}{|\mathcal{T}_{\text{uni}}^b|} \sum_{k \in \mathcal{T}_{\text{uni}}^b} \tv_k^{b}    
    \end{align*}

    \item \textbf{\tam}~\citep{ilharco2023editing}. Applied directly to obtain $\tv_\textrm{uni}$.
   \begin{align*}
        \mathcal{F}(\{ \tv_k^{b} \mid k \in \mathcal{T}_{\text{uni}}^b\}) = \lambda \cdot \frac{1}{|\mathcal{T}_{\text{uni}}^b|} \sum_{k \in \mathcal{T}_{\text{uni}}^b} \tv_k^{b}    
    \end{align*}
    For all experiments with \tam, we use $\lambda = 1.5$. 
    Note that our definition of $\lambda$ for \tam excludes the $1/|\mathcal{T}_{\text{uni}}^b|$ factor (see \Cref{sec:preliminaries}), in contrast to prior work.
    This modification is essential for ensuring reasonable merging performance with \tam, as the number of tasks being fused per block varies throughout the bottom-up fusion process, ranging between $2$ and $M$.

    \item \textbf{\ties}~\citep{NEURIPS2023_1644c9af}. We apply trimming to the full task vectors $\tv_k, k \in [M]$ to retain the top $10\%$ of the parameters before beginning the greedy merging process. 
    This is because selecting the top parameters globally performs better than selecting them within each block.
    During the bottom-up merging process, only the elect sign and disjoint merge steps of \ties are performed. 
    Lastly, we employ the recommended value of $\lambda = 1$.

    \item \textbf{\pcbm}~\citep{du2024parameter}. Applied directly at the block-level by executing the steps of intra-balancing, inter-balancing and drop and rescale at the block-level. In \pcbm, the drop operation for any task vector depends on other task vectors that it is being merged with. This contrasts with \ties where the drop (or trim) operation is solely dependent on the magnitude of values in the task vector. Therefore, global trimming is not possible in \pcbm and we execute it block-wise. We set the sparsity ratio to $10\%$ for vision tasks and $20\%$ for \ac{PEFT} and FFT experiments. Additionally, we employ the recommended value of $\lambda = 1$ across all experiments.

    \item \textbf{\consensusm}~\citep{wang2024localizing}. Applied directly wherein the merging results in not only the unified task vector but also the task specific masks.
    \begin{align*}
        \tv_\textrm{uni}^b, \{\mask_k^b \mid k \in \mathcal{T}_{\text{uni}}^b\} \leftarrow \mathcal{F}(\{ \tv_k^{b} \mid k \in \mathcal{T}_{\text{uni}}^b\})
    \end{align*}
    The task-specific weights for task $k \in \mathcal{T}_{\text{uni}}^b$ corresponding to this merged block $b$ are then reconstructed during the reconstruction process as:
    \begin{align*}
        \bthetah_k^{b} = \btheta_\textrm{pre}^{b} + \tvmtl^b \circ \mask_{k}^b
    \end{align*}
    We use \ties as the algorithm to generate $\tv_\textrm{uni}^b$ within \consensusm.
    Note that the above version of \consensusm corresponds to the compression application in \citep{wang2024localizing}.
    The alternative version of \consensusm that corresponds to merging using masks can be also directly leveraged within \sys.

    \item \textbf{\emr}~\citep{huang2024emrmerging}. Applied directly wherein the merging results in the unified task vector, the task specific masks and task-specific rescalers.
    \begin{align*}
        \tv_\textrm{uni}^b, \{\mask_k^b, \gamma_k^b \mid k \in \mathcal{T}_{\text{uni}}^b\} \leftarrow \mathcal{F}(\{ \tv_k^{b} \mid k \in \mathcal{T}_{\text{uni}}^b\})
    \end{align*}
    The task-specific weights for task $k \in \mathcal{T}_{\text{uni}}^b$ corresponding to this merged block $b$ are then reconstructed during the reconstruction process as:
    \begin{align*}
        \bthetah_k^{b} = \btheta_\textrm{pre}^{b} + \gamma_k^b \cdot \tvmtl^b \circ \mask_{k}^b
    \end{align*}

    \item \textbf{\tsvm}~\citep{gargiulo2025task}. Applied directly per-block as the method originally also operates layerwise. 2D parameters are merged using their task singular vectors while 1D parameters just use averaging for merging as done in their original work.

    \item \textbf{\wudim}~\citep{cheng2025whoever}. Applied directly per-block as the method originally also operates layerwise. We use the same learning rate of $10^{-5}$ as used by the authors within the method. Lastly, we set the number of iterations to $300$ for $8$ tasks and $1000$ for $30$ tasks. 
    
\end{itemize}

\subsection{Vision benchmark}
\label{subsec:appendix_vision_benchmark_details}

The $8$ task vision benchmark~\citep{ilharco2023editing} comprises the following datasets:
\begin{inparaenum}
\item SUN397~\citep{xiao2010sun},
\item Cars~\citep{krause20133d},
\item RESISC45~\citep{cheng2017remote},
\item EuroSAT~\citep{helber2019eurosat},
\item SVHN~\citep{yuval2011reading},
\item GTSRB~\citep{stallkamp2011german},
\item MNIST~\citep{lecun1998mnist}, and
\item DTD~\citep{cimpoi2014describing}.
\end{inparaenum}
We extend the $8$ task vision benchmark with $12$ additional datasets sourced from ~\citep{wang2024localizing}, including:
\begin{inparaenum}
    \setcounter{enumi}{8}
\item CIFAR100 \citep{krizhevsky2009learning},
\item STL10 \citep{stl10},
\item Flowers102 \citep{nilsback2008automated},
\item OxfordIIITPet \citep{parkhi2012cats},
\item PCAM \citep{veeling2018rotation},
\item FER2013 \citep{goodfellow2013challenges},
\item EMNIST \citep{cohen2017emnist},
\item CIFAR10 \citep{krizhevsky2009learning},
\item Food101 \citep{bossard14},
\item FashionMNIST \citep{xiao2017fashion},
\item RenderedSST2 \citep{socher2013recursive,radford2019language} and 
\item KMNIST \citep{clanuwat2018deep}.
\end{inparaenum}
The remaining $10$ datasets for our $30$ task benchmark are sourced from \citep{huang2024emrmerging}, which include:
\begin{inparaenum}
    \setcounter{enumi}{20}
\item Weather~\citep{xiao2021classification},
\item Vegetables~\citep{ahmed2021dcnn},
\item MangoLeafBD~\citep{ahmed2023mangoleafbd},
\item Landscape Recognition~\citep{Landscape},
\item {Beans}~\citep{beansdata},
\item {Intel Images}~\citep{bansal2019intel},
\item Garbage Classification~\citep{cchang_2018}
\item {Kvasir}~\citep{pogorelov2017kvasir},
\item KenyanFood13~\citep{jalal2019scraping} and
\item Dogs~\citep{KhoslaYaoJayadevaprakashFeiFei_FGVC2011}
\end{inparaenum}

\subsection{\Acf{PEFT}}
\label{subsec:appendix_peft_details}

We fine-tune (IA)\(^3\) modules on $11$ diverse datasets, including RTE~\citep{giampiccolo2007third}, CB~\citep{de2019commitmentbank}, Winogrande~\citep{sakaguchi2021winogrande}, WiC~\citep{pilehvar2018wic}, WSC~\citep{levesque2012winograd}, COPA~\citep{roemmele2011choice}, H-SWAG~\citep{zellers2019hellaswag}, Story Cloze~\citep{sharma2018tackling}, and ANLI~\citep{nie2019adversarial} (R1 to R3). 
In addition, we leverage prompt templates from the Public Pool of Prompts (P3)~\citep{bach2022promptsource} which convert each dataset example into a text-to-text format, where each label is mapped to a unique string representation.  
We report the median performance across all templates for each dataset.
Evaluating multiple prompt templates increases the evaluation runtime significantly.  
To ensure the runtime remains manageable, we cap the maximum number of test samples at $1000$ per dataset.

\subsection{Full parameter fine-tuning}
\label{subsec:appendix_full_parameter_ft_details}

The $7$ datasets that we consider for fine-tuning include:
PAWS~\citep{zhang-etal-2019-paws}, QASC~\citep{QASC_Khot_Clark_Guerquin_Jansen_Sabharwal_2020}, QuaRTz~\citep{tafjord-etal-2019-quartz}, Story Cloze~\citep{sharma2018tackling}, WikiQA~\citep{yang-etal-2015-wikiqa}, Winogrande~\citep{sakaguchi2021winogrande}, and WSC~\citep{levesque2012winograd}.
During training and evaluation, we apply a specific prompt template from P3~\citep{bach2022promptsource} to each dataset. Each model is trained for up to \num{75000} optimization steps, with early stopping if validation accuracy does not improve over five consecutive evaluation rounds. Performance is evaluated every \num{5} steps for WSC and every \num{100} steps for other datasets, using the full validation set. Training is conducted using the Adam optimizer with a constant learning rate of \num{0.0001} and an effective batch size of \num{1024}. The maximum sequence length is set to \num{128}, and bfloat16 precision is used for both training and evaluation.

\subsection{Multi-modal benchmark}
\label{subsec:appendix_muttimodal_details}

We evaluate the performance of merging fine-tuned checkpoints of \beit model~\citep{wang2022imageforeignlanguagebeit} under \sys. We consider four datasets: COCO Captioning~\citep{10.1007/978-3-319-10602-1_48} (Image Captioning), ImageNet-1k~\citep{5206848} (Image Classification), NLVR2~\citep{suhr2019corpusreasoningnaturallanguage} (Visual Reasoning) and COCO Retrieval~\citep{10.1007/978-3-319-10602-1_48} (Image-Text Retrieval). The individual checkpoints, each fine-tuned on one of these datasets, are available in UniLM repository\footnote{\url{https://github.com/microsoft/unilm}}. We merge layers that are common to all checkpoints \ie excluding the task-specific classification heads. We report accuracy for ImageNet-1k, NLVR2 and COCO Retrieval, meanwhile for COCO Captioning, we report BLEU-4~\citep{10.3115/1073083.1073135}, CIDEr~\citep{7299087}, ROUGE-L~\citep{lin-2004-rouge} and METEOR~\citep{banerjee-lavie-2005-meteor}.

\section{Additional results}
\label{sec:appendix_results}

In this section, we provide the remaining results for each benchmark.

\subsection{Vision benchmark}
\label{subsec:appendix_vision_exps}

\begin{figure}[h!]
    \centering
    \includegraphics[]{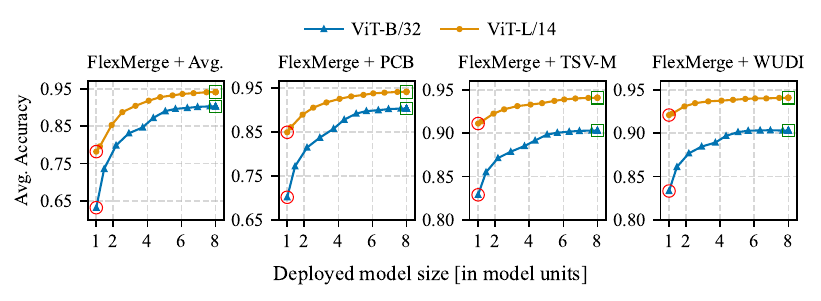}
    \caption{Accuracy-size trade-off for $8$ tasks.}
    \label{fig:appendix_vision}
\end{figure}

\begin{figure}[h!]
    \centering
    \includegraphics[]{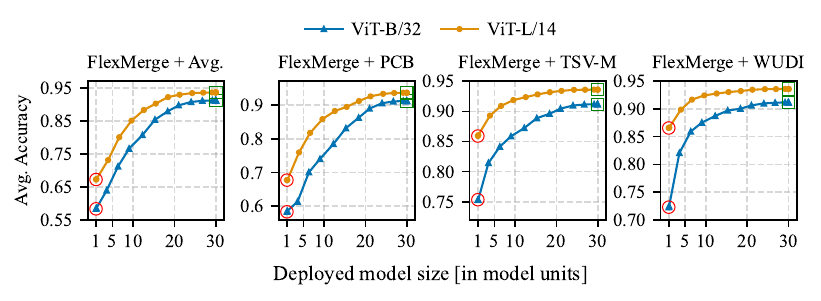}
    \caption{Accuracy-size trade-off for $30$ tasks.}
    \label{fig:appendix_vision2}
\end{figure}

\Cref{fig:appendix_vision,fig:appendix_vision2} show the results on $8$ and $30$ tasks respectively, for Averaging, \pcbm, \tsvm and \wudim extending our results in \Cref{fig:vision}. 
\sys + Avg. gains over $15\%$ in accuracy by just doubling the size from $1\times$ to $2\times$ in the $8$-task benchmark under the \vitb model.
\pcbm starts with higher accuracy at size $1\times$ than vanilla averaging, thanks to its competition balancing procedure.
Yet, it significantly benefits from increasing the deployed model size.
Under the same setup as above, \sys + \pcbm gains $7.5\%$, by just doubling the size from $1\times$ to $2\times$.  
\tsvm and \wudim significantly lead both prior algorithms at size $1\times$, still showing steep improvements of over $4\%$ from size $1\times$ to $2\times$.  
We observe similar steep initial gains for both the algorithms even with $30$ tasks.
Notably, in all cases, we observe that a near fine-tuning accuracy is reached well in advance of the maximum size, around $6\times$ for $8$ tasks and around $23.5\times$ for $30$ tasks.

\subsection{(IA)\(^3\) \ac{PEFT} benchmark}
\label{subsec:appendix_peft_exps}

\begin{figure}[h!]
    \centering
    \includegraphics[]{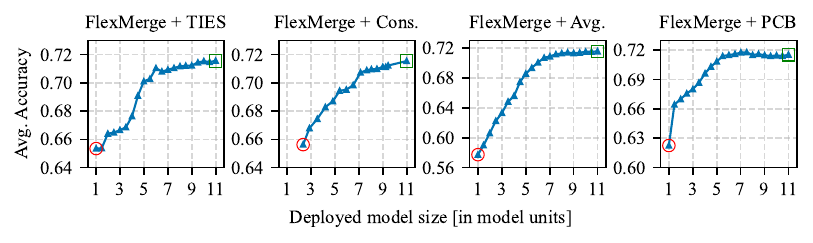}
    \caption{Remaining combinations on the (IA)\(^3\) \ac{PEFT} benchmark with \textbf{\tzerob}.}
    \label{fig:appendix_peft_remaining}
\end{figure}

We extend the results of \Cref{fig:nlp_all} with four additional algorithms in \Cref{fig:appendix_peft_remaining}.
All algorithms demonstrate a steep initial rise in accuracy, except \ties for which the gains appear sharply after a steady initial rise. 
Notably, at size $7\times$, all algorithms are within $1\%$ of the fine-tuning accuracy (denoted by $\square$).
Therefore, practitioners aiming to deploy all $11$ fine-tuned models for high accuracy can be benefit from a reduction of $4$ model units without losing too much accuracy. 

\begin{figure}[h!]
    \centering
    \includegraphics[]{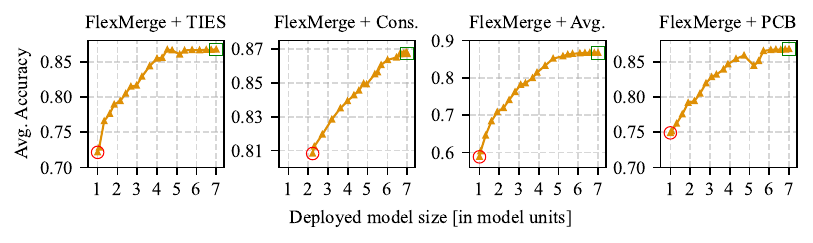}
    \caption{Remaining combinations on the FFT benchmark with \textbf{\tfivelarge}.}
    \label{fig:appendix_t5_large}
\end{figure}

\begin{figure}[h!]
    \centering
    \includegraphics[]{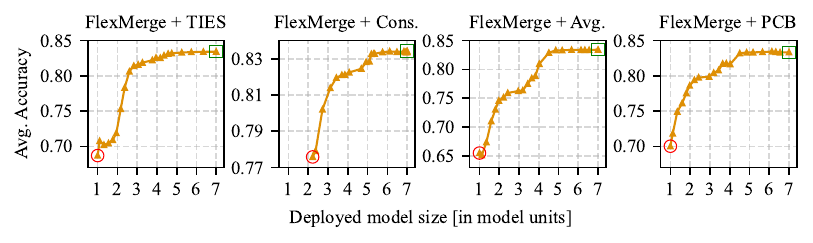}
    \caption{Remaining combinations on the FFT benchmark with \textbf{\tfivebase}.}
    \label{fig:appendix_t5_base}
\end{figure}

\subsection{Full-Parameter Fine-Tuning (FFT) benchmark}
\label{subsec:appendix_fft_exps}
\begin{figure}[t!]
    \centering
    \includegraphics[]{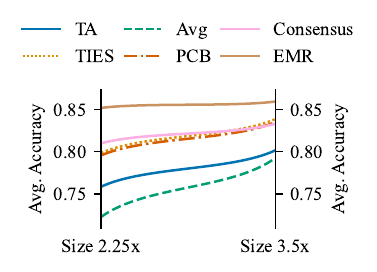}
    \caption{Results on \tfivelarge. Algorithms rankings are not consistent across sizes. In particular, \pcbm and \ties surpass the accuracy of masking-based approach \consensusm while averaging almost attains the accuracy of \tam at size $3.5\times$.}
    \label{fig:algorithm_rankings_t5large}
\end{figure}

We extend the results of \Cref{fig:nlp_all} with four additional merging algorithms on two models. \Cref{fig:appendix_t5_large} and \Cref{fig:appendix_t5_base} chart the results for \tfivelarge and \tfivebase respectively.
Just doubling the deployed model size from $1\times$ to $2\times$ significantly improves the accuracy across all combinations.
We note gains of more than $7\%$, $12\%$ and $4.3\%$ for \ties, Averaging and \pcbm respectively under the \tfivelarge model.
Similarly, we note gains of nearly $9\%$ for both Averaging and \pcbm under the \tfivebase model. 
\ties under the \tfivebase model demonstrates a sharp rise around size $2\times$, reaching $80.6\%$ accuracy at just $2.7\times$ from an accuracy of $68.7\%$ at size $1\times$, a gain of nearly $12\%$.
Except for \sys + \consensusm under \tfivelarge which demonstrates a linear trade-off, all other combinations exhibit a highly favorable trade-off, validating the benefits of our approach across diverse scenarios.
Lastly, we also observe that algorithm rankings remain inconsistent on the \ac{FFT} benchmark as well when sizes are increased, as illustrated in \Cref{fig:algorithm_rankings_t5large}.

\subsection{Multi-modal benchmark}
\label{subsec:appendix_multimodal_exps}

\begin{figure}[h!]
    \centering
    \includegraphics[]{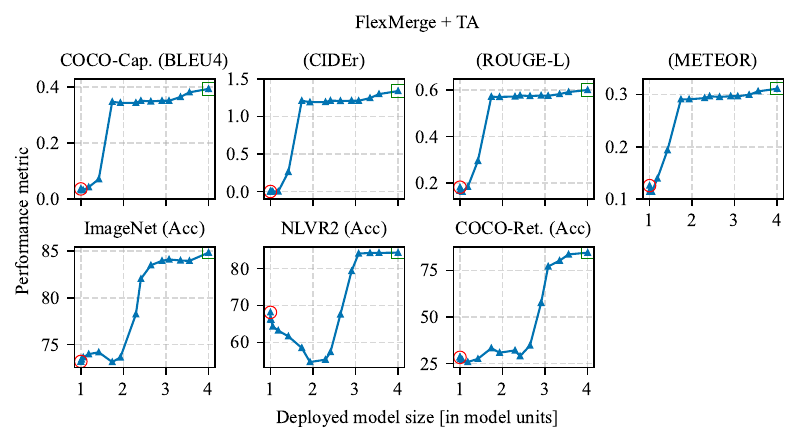}
    \caption{Accuracy-size trade-off for \sys + \tam on the multi-modal benchmark with \textbf{\beit}. 
    The performance metrics for each dataset are shown in parentheses. }
    \label{fig:multimodal_ta}
\end{figure}

\begin{figure}[h!]
    \centering
    \includegraphics[]{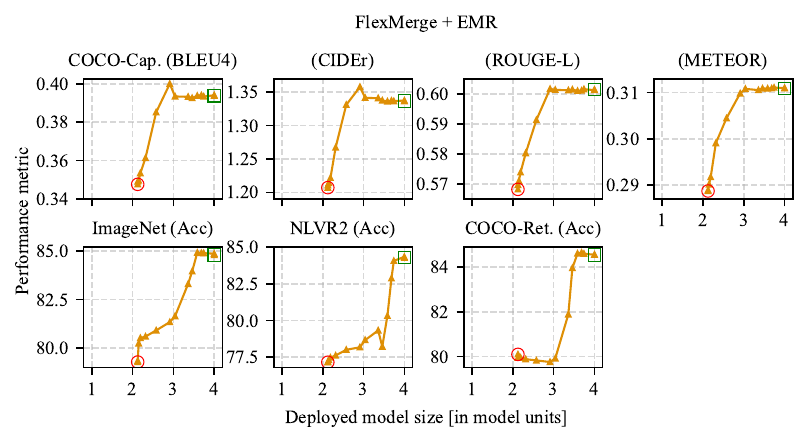}
    \caption{Accuracy-size trade-off for \sys + \emr on the multi-modal benchmark with \textbf{\beit}. The performance metrics for each dataset are shown in parentheses.}
    \label{fig:multimodal_emr}
\end{figure}

In this section, we present the accuracy-size trade-off for the multi-modal benchmark detailed in \Cref{subsec:appendix_muttimodal_details}.
\Cref{fig:multimodal_ta,fig:multimodal_emr} chart the results for \sys + \tam and \sys + \emr respectively.
The performance metrics differ across datasets, therefore we show each dataset separately.
The first row in each figure corresponds to the COCO-Captioning dataset with $4$ metrics while the second row corresponds to ImageNet, NLVR2 and COCO-Retrieval datasets, each evaluated with accuracy. For all metrics, higher values indicate better performance.

Across all datasets, we observe a significant gap between the performance at lowest merged size and the fine-tuned model performance. 
\sys + \tam shows significant benefit of increased size on the COCO-Captioning dataset where the performance steeply grows across all four metrics between size $1\times$ and $2\times$.
For ImageNet, NLVR2 and COCO-Retrieval the performance improves sharply between size $2\times$ and $3\times$.
Interestingly, NLVR2 shows a drop in performance before rising sharply, indicating that merging with related tasks can provide complementary benefits.
\emr has a lowest size of over $2\times$ due to the cost of storing the pre-trained model and the masks.
However, this provides a significantly higher starting performance than \tam.
The performance improvements for \sys + \emr are similar to \sys + \tam, where COCO-Captioning demonstrates a sharp rise between $2\times$ and $3\times$ while the remaining datasets demonstrate a sharp rise between $3\times$ and $4\times$.
Thus, larger sizes can confer beneficial improvements across both algorithms, obviating the need to deploy all fine-tuned models for high performance. 

\subsection{\sys vs \channelm}
\label{subsec:appendix_flex_vs_channel}

\begin{wrapfigure}[18]{r}{0.35\textwidth} %
	\centering
	\vspace{-21 pt}
	\includegraphics[]{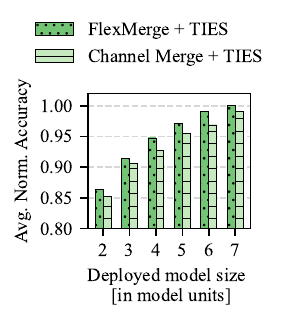}
	\caption{\sys achieves higher accuracy than \channelm across all sizes.}
    \label{fig:appendix_baseline_with_ties}
\end{wrapfigure}
\begin{figure}[h!]
    \centering
    \includegraphics[]{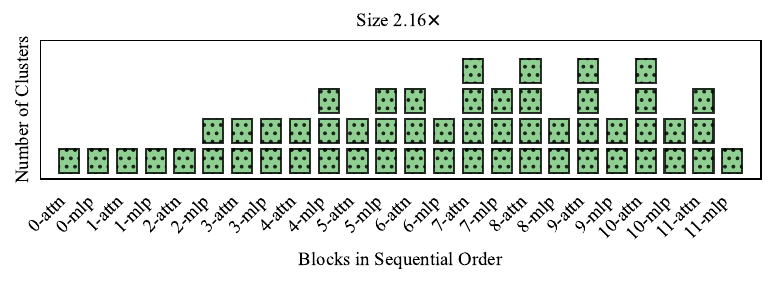}
    \caption{Visualizing the number of clusters per block in \sys model of size $2.16\times$. The number of clusters vary significantly across blocks, ranging from $1$ to $4$. In contrast, for a model of size $2\times$, \channelm will have exactly $2$ clusters in each block. This flexibility enables greater accuracy for \sys over \channelm.}
    \label{fig:visualizing_clusters_size2x}
\end{figure}

In this section, we compare \sys with \channelm under the \ties algorithm, following the same setup as our experiments with \tam in \Cref{fig:baseline_plus_switching}. As shown in \Cref{fig:appendix_baseline_with_ties}, \sys consistently achieves significantly higher accuracy than \channelm across all sizes.
This is because \channelm enforces a fixed number of clusters per block, whereas \sys allows vastly different number of clusters per layer by design.
This flexibility, enabled by iterative pairwise merging, leads to substantial accuracy improvements.
To better understand this effect in \sys, we visualize the number of clusters per block in \Cref{fig:visualizing_clusters_size2x}, corresponding to merged model sizes of $2.16\times$. For clarity, we focus on Attention and MLP blocks, which constitute the bulk of the model size, omitting smaller blocks such as layer norms.
In \Cref{fig:visualizing_clusters_size2x}, we observe that the number of clusters per block varies significantly—from 1 to 4—whereas \channelm, at a similar model size, would enforce exactly 2 clusters per block, limiting its effectiveness. 

\subsection{\ac{OOD} performance of \sys}
\label{subsec:appendix_OOD}

To assess the \ac{OOD} performance of \sys, we conducted additional experiments on 6 OOD tasks using models merged from 8 in-domain tasks. For each OOD task, the prediction is obtained by ensembling the outputs of the 8 in-domain branches of the merged model (see \Cref{fig:intro_fig}(a)). 

\textbf{Setup:} In-domain tasks are SVHN, Cars, RESISC45, EuroSAT, SUN397, GTSRB, MNIST and DTD. Out-of-domain tasks are Weather, PCAM, Flowers102, Landscapes, Beans and Food101. We consider TIES as the merging algorithm as it known for its good performance OOD. 

\textbf{Results.} \Cref{tab:ood_perf1} presents the average accuracy of each in-domain expert on the 6 OOD tasks. \Cref{tab:ood_perf2} presents the average accuracy of \sys's merged model for varying sizes on the 6 OOD tasks. We observe that across all model sizes $> 1\times$, \sys consistently outperforms the best single in-domain expert on these OOD tasks, suggesting that our method not only preserves but can also improve OOD generalization.

\begin{table}[h]
    \centering
    \caption{Average OOD Accuracy of Individual In-domain Experts. Bold-faced values indicate the best in-domain expert.}
    \resizebox{\linewidth}{!}{
    \begin{tabular}{ccccccccc}
        \toprule
       Model $\downarrow$ $\mid $ Dataset $\rightarrow$ & SVHN & Cars & RESISC45 & EuroSAT & SUN397 & GTSRB & MNIST & DTD\\
       \midrule
       ViT-B-32  & 0.5538 & 0.6230 & 0.6268 & 0.5768 & \textbf{0.6366} & 0.5083 & 0.5060 & 0.5910 \\
       ViT-L-14 & 0.7259 & 0.7152 & 0.6908 & 0.6797 & 0.7096 & \textbf{0.7328} & 0.7298 & 0.7193 \\
       \bottomrule
    \end{tabular}}
    \label{tab:ood_perf1}
\end{table}

\begin{table}[h]
    \centering
    \caption{Average OOD Accuracy of FlexMerge at Varying Sizes. Bold-faced values indicate higher accuracy than the best in-domain expert in \Cref{tab:ood_perf1}.}
    \resizebox{\linewidth}{!}{
    \begin{tabular}{ccccccccc}
        \toprule
       Model $\downarrow$ $|$ Merged Size $\rightarrow$ & 1x & 1.5x & 2x & 2.5x & 3x & 4x & 6x & 8x\\
       \midrule
       ViT-B-32  & 0.6289 & \textbf{0.6410} & \textbf{0.6475} & \textbf{0.6539} & \textbf{0.6567} & \textbf{0.6586} & \textbf{0.6548} & \textbf{0.6595} \\
       ViT-L-14 & \textbf{0.7390} & \textbf{0.7394} & \textbf{0.7398} & \textbf{0.7403} & \textbf{0.7416} & \textbf{0.7453} & \textbf{0.7464} & \textbf{0.7488} \\
       \bottomrule
    \end{tabular}}
    \label{tab:ood_perf2}
\end{table}

\subsection{Scaling laws for flexible model merging}
\label{subsec:appendix_scaling_laws}

\begin{figure}[h!]
    \centering
    \begin{subfigure}[b]{0.32\textwidth}
        \includegraphics[width=\linewidth]{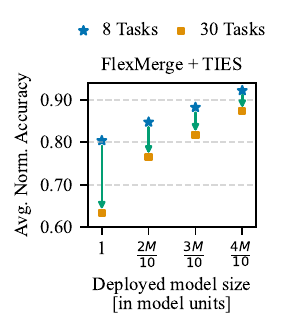} %
    \end{subfigure}
    \hfill
    \begin{subfigure}[b]{0.32\textwidth} %
        \includegraphics[width=\linewidth]{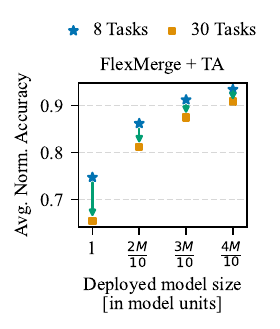} %
    \end{subfigure}
    \hfill
    \begin{subfigure}[b]{0.32\textwidth}
        \includegraphics[width=\linewidth]{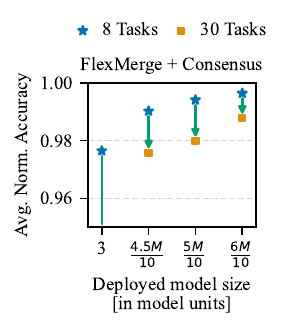} %
    \end{subfigure}

    \caption{Deploying model sizes in proportion to the number of tasks ($M$) scales better than fixed size models ($1\times$ or $3\times$) as shown.}
    \label{fig:appendix_scaling_law}
\end{figure}

As the number of tasks increases, maintaining high accuracy in the merged model becomes challenging. 
While newer methods enhance accuracy scaling, they encounter limitations when the model size is fixed. 
Thanks to flexible model merging, this scaling can be significantly improved if \textit{the deployed model size is chosen in proportion to the number of tasks}.
To illustrate this, we chart in \Cref{fig:appendix_scaling_law}, the drop in average normalized accuracy from $8$ tasks to $30$ tasks, when deploying a model of fixed size $1\times$ vs sizes chosen proportionally to $M$.
Noticeably, the drop becomes smaller as the proportions increase, becoming $< 5\%$ at size $\nicefrac{4M}{10}$ from $> 16.5\%$ at size of $1\times$ for \ties. 

We observe similar results across other algorithms.
For \tam at size $1\times$, the average normalized accuracy drops by $9.2\%$ when scaling from 8 to 30 tasks.
With \sys, this degradation can be significantly mitigated by adjusting the deployed model size in proportion to the number of tasks $(M)$. 
For instance, setting the deployed model size to $\nicefrac{3M}{10}$ reduces the drop to $3.7\%$, and increasing it to $\nicefrac{4M}{10}$ brings the drop down further to $2.5\%$.
Even for advanced algorithms such as \consensusm, fixed-size models can suffer substantial degradation. 
At a fixed size of $3\times$, the accuracy drop from 8 to 30 tasks for \consensusm is $14.5\%$ (beyond the y-axis limits of the figure). However, increasing the deployed model size to $\nicefrac{4.5M}{10}$ reduces the drop to $2.5\%$, and to less than $1\%$ at $\nicefrac{6M}{10}$.
While fixed-size models struggle to scale effectively with the number of tasks, our results highlight the need to rethink accuracy scaling—advocating for model sizes that grow proportionally with task count.

\subsection{Ablations of the merging procedure}
\label{subsec:appendix_ablations}
\begin{figure}[t!]
    \centering
    \includegraphics[]{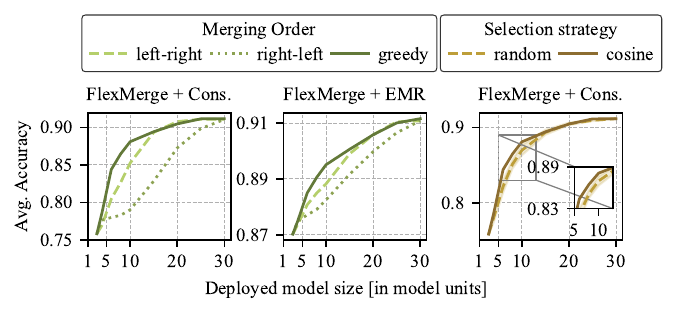}
    \caption{Greedy emerges as the best over left-to-right and right-to-left for both \sys + \consensusm and \sys + \emr whereas cosine performs better than random.}
    \label{fig:appendix_merging_order}
\end{figure}

In this section, we include the remaining ablation results.
In particular, we explore the impact of merging order on \consensusm and \emr.
We consider three merging orders: left-to-right, right-to-left and greedy.
In the left-to-right order, we execute pairwise iterative merging using cosine similarity in block 1, followed by block 2, block 3 and so on (see \Cref{fig:intro_fig}).
Hence, all task-blocks in block $i$ are fully merged before carrying out any merging in block $i+1$.
The right-to-left merging does exactly the same, but in reverse order, starting from the last block.
In contrast to both, greedy merging can select any pairs in any block depending upon their cosine similarity, without any restriction on the order.
\Cref{fig:appendix_merging_order} shows the results for \sys + \consensusm and \sys + \emr.
Similar to our results in \Cref{fig:merging_ablations}, right-to-left performs the worst.
As final layers tend to be more specialized, merging them first significantly hurts accuracy. 
Intuitively, left-to-right merging seems ideal. However, strictly following this order proves too rigid, and a more flexible greedy merging approach performs best.
We further analyze the ordering in greedy in \Cref{subsec:merging_order_analysis}.

\begin{figure}[t!]
    \centering
    \includegraphics[]{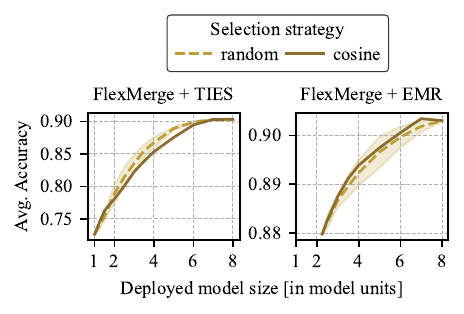}
    \caption{For \ties, its trimming component makes similarity comparisons noisy. Hence, random selection performs slightly better than cosine. In contrast, cosine similarity performs better on \emr and most other cases (see also \Cref{fig:merging_ablations,fig:appendix_merging_order}).}

    \label{fig:appendix_selection_strategy}
\end{figure}

We also explore the impact of random selection over cosine similarity based selection on more merging algorithms, namely \ties and \emr, extending our results in \Cref{fig:merging_ablations}.
In \Cref{fig:appendix_merging_order}, cosine performs better than random on \sys + \consensusm.
However in \Cref{fig:appendix_selection_strategy}, random selection shows slightly better performance for \sys + \ties. 
This is likely due to the trim component of \ties which makes similarity comparisons noisy.
As detailed in \Cref{subsubsec:how_to_apply_at_blocklevel}, we retain only the top $10\%$ parameters in the full task vectors before beginning the bottom-up merging process when using \ties.
While sparsification enables better merging due to reduced interference~\cite{NEURIPS2023_1644c9af}, it also renders similarity comparisons noisy, leading to a slight drop in performance compared to random selection.
With \sys + \emr, random shows high variance, and once again cosine performs better on average than random.
In summary, cosine is generally superior across most cases in \Cref{fig:merging_ablations,fig:appendix_merging_order,fig:appendix_selection_strategy}.

\subsection{Reconstruction latency of masking based approaches}
\label{subsec:appendix_reconstruction_latency}

\subsubsection{How do we measure the reconstruction latency?}
\label{subsec:how_to_measure_latency}

Approaches such as \consensusm and \emr merge into a unified task vector $\tvmtl$ and task-specific masks $(\mask_1, \ldots, \mask_M)$. 
To reconstruct the task-specific model, they also store $\btheta_\textrm{pre}$ and reconstruct as follows:
\begin{equation}
	\bthetah_t = \btheta_\textrm{pre} + \tvmtl \circ \mask_t
\end{equation}
Under \sys, the size of the deployed model varies.
Consider for a certain task $t$, two sets -- $B_t^F$ and $B \backslash B_t^F$, comprising the set of all blocks that were fused and retained as original respectively.
Notice the subscript $t$ which indicates that these sets could be different for different tasks depending on how the greedy fusion occurred.
For each fused block $b \in B^F_t$, let $\mathcal{T}^b$ denote the subset of tasks which got fused.
Then \sys will store the corresponding $\tvmtl^b$ along with the task specific masks $\{\mask_k^b\}_{k \in \mathcal{T}^b}$ and $\btheta_\textrm{pre}^b$.
Under \sys, we then reconstruct as follows:
\begin{equation}
	\label{eqn: reconstruction_fused_blocks}
	\text{for } b \in B^F_t:  \bthetah_t^b = \btheta_\textrm{pre}^b + \tvmtl^b \circ \mask_{t}^b
\end{equation}
And, no reconstruction is required for unfused blocks:
\begin{equation}
\text{for } b \in B \backslash B^F_t: \bthetah_t^b = \btheta_t^b
\end{equation}
The reconstruction latency therefore depends upon the size of $B^F_t$.
Ideally, we would measure the reconstruction latency as the time required to execute  \cref{eqn: reconstruction_fused_blocks} averaged across all tasks $t \in [M]$.
However, this would mean that the reconstructed parameters $\bthetah_t$ occupy additional storage alongside $\btheta_\textrm{pre}$.
To restrict this extra storage, we assume that the reconstruction happens in-place on $\btheta_\textrm{pre}$ as follows:
\begin{equation}
	\label{eqn: reconstruction_fused_blocks_1}
	\text{for } b \in B^F_t:  \btheta_\textrm{pre}^b = \btheta_\textrm{pre}^b + \tvmtl^b \circ \mask_{t}^b
\end{equation}
Once the inference for task $t$ is completed, the pre-trained parameters are restored back to be ready for the next reconstruction:
\begin{equation}
	\label{eqn: reconstruction_fused_blocks_2}
	\text{for } b \in B^F_t:  \btheta_\textrm{pre}^b = \btheta_\textrm{pre}^b - \tvmtl^b \circ \mask_{t}^b
\end{equation}
Thus, we report the reconstruction latency to be the total time required to execute both \Cref{eqn: reconstruction_fused_blocks_1} and \Cref{eqn: reconstruction_fused_blocks_2}, averaged across all tasks.

\subsubsection{How does increasing the deployed model size lower the reconstruction latency?}
As noted earlier, the reconstruction latency depends on the size of $B^F_t$.
As the deployed model size progressively increases, more and more blocks move from $B^F_t$ to $B \backslash B^F_t$, and the time to reconstruct consequently reduces.
The lowest deployed size \ie where all blocks are fused, $B^F_t = B$, incurs the highest reconstruction overhead.
Conversely, the maximum possible deployed size (of $M\times$), with $B^F_t = \emptyset$, incurs zero reconstruction latency.
Thus by employing larger deployed models generated by \sys, practitioners can effectively reduce the reconstruction overhead in time-critical applications.

\begin{figure}[h!]
    \centering
    \includegraphics[]{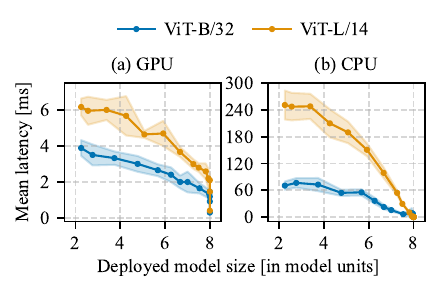}
    \caption{\sys enables lowering of the reconstruction latency overheads for approaches such as \consensusm~\cite{wang2024localizing} and \emr~\cite{huang2024emrmerging} by offering flexible control over the deployed model size. The default overhead at the lowest size can be significant, exceeding 50\% for the forward pass time, which is \SI{12}{\milli\second} for \vitl and \SI{7}{\milli\second} for \vitb on GPU.}
    \label{fig:masking_consensus}
\end{figure}
\textbf{Results.} \Cref{fig:masking_consensus} shows the average reconstruction latency for \consensusm, measured on both CPU and GPU. The latency for \emr closely matches that of \consensusm, as the additional scalar multiplications in \emr incur negligible overhead; the dominant cost arises from the application of masks.
As illustrated in \Cref{fig:masking_consensus}, merging into larger models using \sys can significantly reduce reconstruction latency. For ViT-L/14 on CPU, latency drops from \SI{240}{\milli\second} to \SI{0}{\milli\second}, and for ViT-B/32, from \SI{60}{\milli\second} to \SI{0}{\milli\second}, depending on the deployed model size. On GPU, the latency similarly decreases—from \SI{6}{\milli\second} to \SI{0}{\milli\second} for \vitl and from \SI{4}{\milli\second} to \SI{0}{\milli\second} for \vitb.
To put this in perspective, the forward pass takes \SI{12}{\milli\second} for \vitl and \SI{7}{\milli\second} for \vitb. On GPU, reconstruction latency can exceed 50\% of the forward pass time, while on CPU, the overhead is several times higher. These reconstruction costs can be significant in time-critical scenarios.
By enabling fine-grained control over the deployed model size, \sys allows \consensusm and \emr to achieve not only higher accuracy but also lower reconstruction latency.

\subsection{Merging order analysis}
\label{subsec:merging_order_analysis}

\begin{figure*}[h!]
    \centering
    \includegraphics[width=\textwidth]{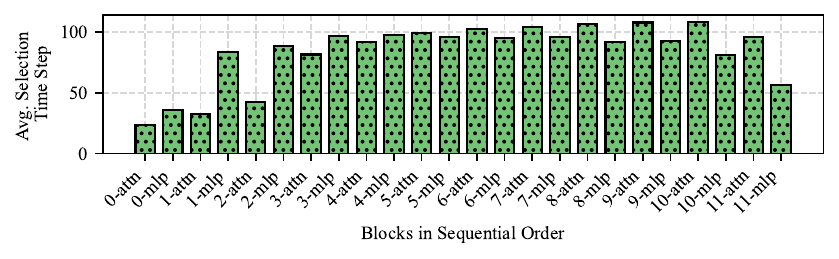}
    \caption{Visualizing the order of greedy selection blocks for \sys + \tam on $8$ task vision benchmark with \vitb.
    We focus only on Attention and MLP blocks as they share the biggest portion of the size of the model.
    We observe that the initial and final blocks get selected ahead of the intermediate blocks on average, indicated by the smaller values of their average selection time step.
    }
    \label{fig:merging_analysis}
\end{figure*}

\Cref{fig:merging_analysis} visualizes the order of greedy block selection for the run corresponding to \sys + \tam on $8$ task vision benchmark with \vitb (shown in \Cref{fig:vision}).
We focus only on Attention and MLP layers as they share the biggest portion of the size of the model.
There are $12$ transformer layers in the \vitb model, resulting in a total of $24$ blocks that we consider in the sequential order. 
The number of possible merges total to $(M-1) \times 24 = 7 \times 24 = 168$, with at most $(M-1)$ merges possible for each of the $24$ blocks.
The time steps are assigned in order of selection starting from $0$ to $167$.
We report the average selection time step for every block in \Cref{fig:merging_analysis}.
We observe that the initial and final blocks are, on average, selected earlier than the intermediate blocks, as indicated by their smaller average selection time step. 
While the overall trend aligns with the common understanding that earlier layers in neural networks learn general features shared across tasks and later layers capture more task-specific features with lower correlation, we find that the final few blocks may also exhibit higher cosine similarity and be selected early for merging. 
This flexibility, enabled by \sys's greedy selection process, facilitates the efficient trade-off between model size and accuracy.

\begin{figure}[h!]
    \centering
    \includegraphics[]{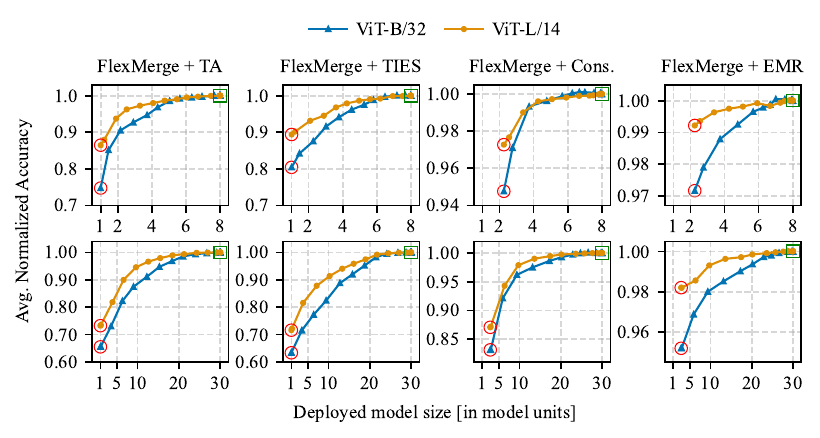}
    \caption{Vision benchmark results (\Cref{fig:vision}) shown with average normalized accuracy.}
    \label{fig:vision_normalized}
\end{figure}

\begin{figure}[h!]
    \centering
    \includegraphics[]{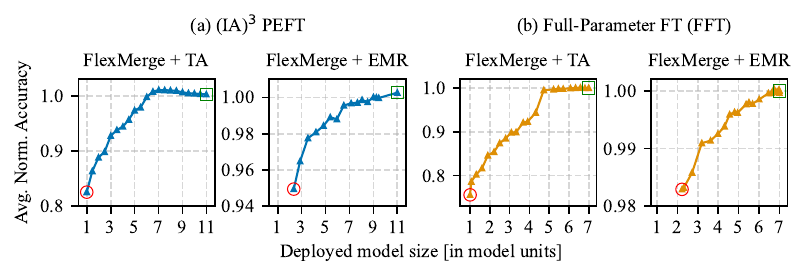}
    \caption{\ac{PEFT} and FFT benchmark results (\Cref{fig:nlp_all}) shown with average normalized accuracy.}
    \label{fig:nlp_all_normalized}
\end{figure}

\subsection{Main results presented with normalized accuracy}
\label{subsec:appendix_normalized_accuracy}

We presented results using average accuracy in \Cref{fig:vision,fig:nlp_all}. For completeness, \Cref{fig:vision_normalized,fig:nlp_all_normalized} show the corresponding results using average normalized accuracy, computed by dividing the accuracy of the merged model by the fine-tuning accuracy on each task, and then averaged across tasks.

\subsection{Dataset-wise results}
\label{subsec:appendix_datasetwise_results}

\begin{figure}[t!]
    \centering
    \includegraphics[]{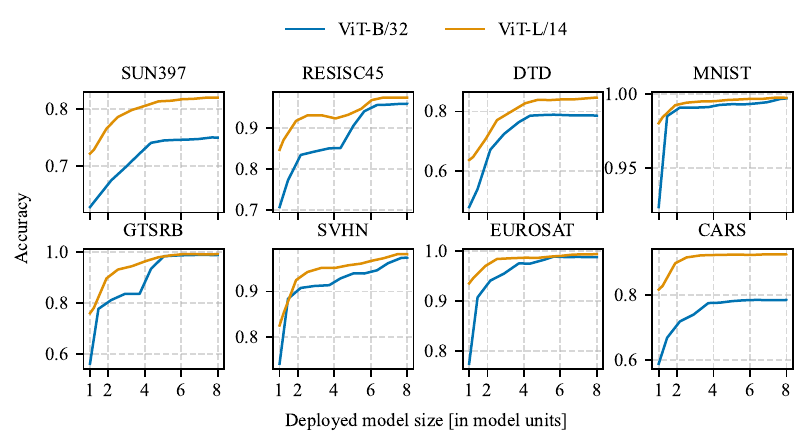}
    \caption{Dataset-wise results for \sys + \tam on the vision benchmark (\Cref{fig:vision}).}
    \label{fig:vision_datasetwise}
\end{figure}

\begin{figure}[t!]
    \centering
    \includegraphics[]{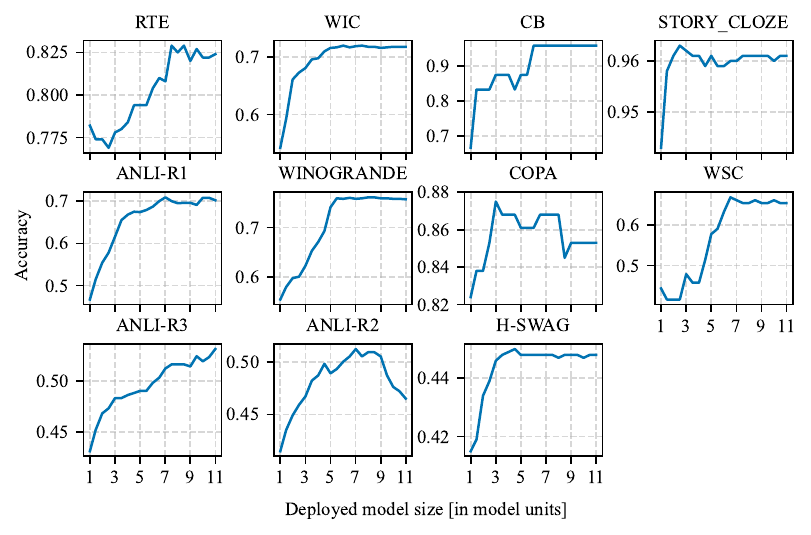}
    \caption{Dataset-wise results for \sys + \tam on the \ac{PEFT} benchmark (\Cref{fig:nlp_all}(a)).}
    \label{fig:peft_datasetwise}
\end{figure}

\begin{figure}[t!]
    \centering
    \includegraphics[]{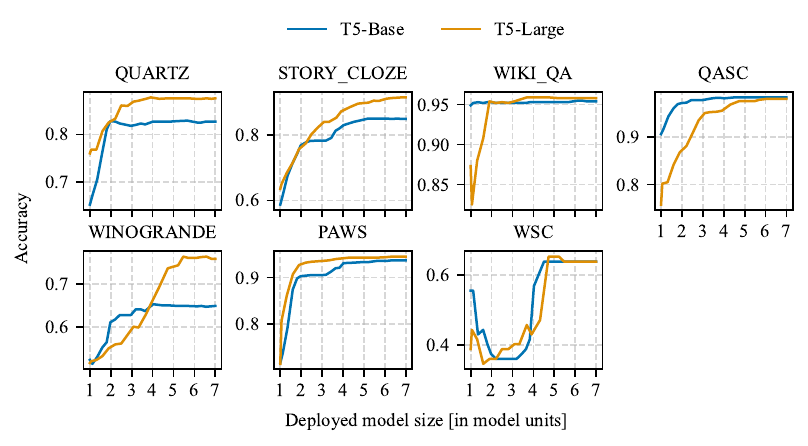}
    \caption{Dataset-wise results for \sys + \tam on the FFT benchmark (\Cref{fig:nlp_all}(b)).}
    \label{fig:fft_datasetwise}
\end{figure}

\Cref{fig:vision_datasetwise,fig:peft_datasetwise,fig:fft_datasetwise} chart the dataset-wise results for \sys + \tam, corresponding to the plots in \Cref{fig:vision,fig:nlp_all}.

 {
\color{blue}

 }
\section{Compute Resources}
\label{sec:compute_resources}

All experiments were conducted on an internal compute cluster comprising 2 x AMD EPYC 7543 32-Core 2.8GHz CPU Processor, equipped with 8 x NVIDIA A100 SXM4 80GB GPU. 
All of our experiments individually use only 1 out of the 8 GPU units.
While the merging itself is efficient, the evaluation of test accuracy consumes bulk of the time and compute.
The wall-clock time can range anywhere between \SI{3}{\hour} for $8$ tasks to up to \SI{12}{\hour} for $30$ tasks under the \vitl model for evaluating up to $20$ merged sizes in the size range $[1, 30]$. Similarly, the wall clock times range up to \SI{2}{\hour} for full-parameter fine-tuning, up to \SI{8}{\hour} for \ac{PEFT} and up to \SI{5}{\hour} for multi-modal test evaluations.
Across all experiments presented in this article, we estimate the total virtual CPU and GPU time to be approximately \SI{1200}{\hour} each.

\section{LLM Usage Statement}
\label{sec:llm_usage}

We acknowledge the use of LLMs in this work, limited to coding assistance, identifying potentially relevant related work, and improving the clarity and grammar of the manuscript. All LLM-generated content was reviewed and verified by the authors.

\end{document}